\documentclass[final,3p,times,twocolumn]{elsarticle}

\usepackage{lineno}
\modulolinenumbers[5]

\journal{Journal of .....}
\usepackage{graphicx}
\usepackage{dcolumn}
\usepackage{bm}
\usepackage{multirow} 
\usepackage{color}
\usepackage{graphicx} 
\usepackage{mathtools}
\usepackage{multirow}
\usepackage{caption}
\usepackage{amsmath} 
\usepackage{chemformula} 
\usepackage{csquotes} 
\biboptions{numbers,sort&compress}
\usepackage{placeins}
\usepackage{cancel}
\usepackage[normalem]{ulem}
\usepackage{siunitx}
\usepackage{bm}
\usepackage[capitalise]{cleveref}
\usepackage{xr}
\usepackage{comment}
\usepackage{csvsimple}
\usepackage{breqn}
\usepackage{stfloats}
\usepackage{braket}
\usepackage{graphicx}
\usepackage{subcaption}

\newcommand{\eref}[1]{Eq.~(\ref{#1})}

\newcommand{\sref}[1]{Section~\ref{#1}}

\makeatletter
\newcommand*{\addFileDependency}[1]{
  \typeout{(#1)}
  \@addtofilelist{#1}
  \IfFileExists{#1}{}{\typeout{No file #1.}}
}
\makeatother

\usepackage{amsmath, amssymb}
\usepackage{algorithm}
\usepackage{algpseudocode}
\usepackage{tcolorbox}
\usepackage{array}

\begin{document}

\begin{frontmatter}
\title{Stabilized Maximum-Likelihood Iterative Quantum Amplitude Estimation for Structural CVaR under Correlated Random Fields}


\author[1,b]{Alireza Tabarraei\corref{mycorrespondingauthor}}
\cortext[mycorrespondingauthor]{Corresponding author}
\ead{atabarra@charlotte.edu}

\address[1]{Department of Mechanical Engineering and Engineering Science, The University of North Carolina at Charlotte, Charlotte, NC 28223, USA}
\address[b]{School of Data Science, The University of North Carolina at Charlotte, Charlotte, NC 28223, USA}

\begin{abstract}
Conditional Value-at-Risk (CVaR) is a fundamental tail-risk measure in stochastic structural mechanics, yet its accurate evaluation under high-dimensional, spatially correlated material uncertainty remains computationally prohibitive for classical Monte Carlo methods. Leveraging bounded-expectation reformulations of CVaR compatible with quantum amplitude estimation, we develop a quantum-enhanced inference framework that casts CVaR evaluation as a statistically consistent, confidence-constrained maximum-likelihood amplitude estimation problem.
The proposed approach extends iterative quantum amplitude estimation (IQAE) by embedding explicit maximum-likelihood inference within a rigorously controlled interval-tracking architecture. To ensure global correctness under finite-shot noise and the non-injective oscillatory response induced by Grover amplification, we introduce a stabilized inference scheme incorporating multi-hypothesis feasibility tracking, periodic low-depth disambiguation, and a bounded restart mechanism governed by an explicit failure-probability budget. This hybrid likelihood--interval formulation preserves the quadratic oracle-complexity advantage of amplitude estimation while providing provable finite-sample confidence guarantees and substantially reduced estimator variance.
The framework is demonstrated on 
examples with spatially correlated lognormal Young’s modulus fields generated via a Nyström low-rank Gaussian kernel model. Numerical experiments show that the proposed estimator achieves significantly lower oracle complexity than classical Monte Carlo CVaR estimation at comparable confidence levels, while maintaining rigorous global statistical reliability.
This work establishes a practically robust and theoretically grounded quantum-enhanced methodology for tail-risk quantification in stochastic continuum mechanics and introduces a new class of confidence-constrained maximum-likelihood amplitude estimators for engineering uncertainty quantification.
\end{abstract}

\begin{keyword}
\texttt Iterative quantum amplitude estimation \sep Conditional value-at-risk (CVaR) \sep structural reliability \sep quantum computing \sep Finite element analysis.
{\MSC[2010] 00-01\sep  99-00}
\end{keyword}
\end{frontmatter}

\section{Introduction}

In computational mechanics, the quantification of structural response under uncertainty has become increasingly important
as engineers seek to design systems that are not only efficient under nominal conditions but also resilient to
variability in material properties, loads, and boundary conditions
\cite{Ghanem1991KL,LeMa2010StochasticFEM,Xiu2010UQBook,UQforEngineers2012}.
Traditional uncertainty quantification approaches in engineering have primarily focused on mean responses, variances,
and failure probabilities derived from limit-state functions and reliability indices
\cite{Melchers1999Reliability,Madsen2006Methods,DerKiureghian2005Reliability}.
While these measures are effective for characterizing average behavior and failure likelihood, they are often
insufficient for assessing the severity of rare but consequential structural responses.

In many engineering applications, safety margins, serviceability limits, and certification decisions are governed by
extreme responses rather than by central tendencies.
Unusually large compliance, stress, displacement, or fatigue responses can dominate structural performance and lead to
catastrophic failure, even when mean behavior remains acceptable
\cite{Nowak2001Reliability,Ang2013Probability}.
From a physical perspective, such extremes are frequently driven by localized failure mechanisms, including shear bands,
cracks, and strain-softening zones, which generate highly concentrated stress fields and dominate the upper tail of the
response distribution.
These phenomena arise in impact, fracture, and failure of heterogeneous materials and have been extensively studied in
computational mechanics
\cite{Bazant1988CrackBand,elapolu2021mechanical,Simo1992SDM,Ortiz1999Discontinuous,tabarraei2013two}.
Because these localized responses are rare but severe, they are poorly characterized by mean or variance alone,
underscoring the need for tail-sensitive risk measures.

Among such measures, value-at-risk (VaR) and conditional value-at-risk (CVaR) have been extensively studied in finance,
operations research, and risk management \cite{rockafellar2000optimization,Acerbi2002CVaR,rockafellar2002conditional,Pflug2000VaR}.
CVaR is particularly attractive due to its coherent and convex formulation, which quantifies tail severity by averaging
responses beyond a prescribed quantile threshold.
These properties have motivated growing interest in engineering and structural mechanics, where CVaR offers a principled
compromise between robustness and tractability \cite{lee2023bi}.
In contrast to worst-case formulations, which are often overly conservative and numerically unstable, CVaR preserves
sensitivity to extreme responses while remaining amenable to analysis and optimization
\cite{rockafellar2002conditional,UQforEngineers2012}.
Recent work has explored CVaR-based formulations for reliability assessment and risk-aware design under correlated
uncertainty in structural systems, including surrogate-based strategies for estimating tail risk in nonlinear finite
element models \cite{lee2023bi,Airaudo2023UseCVaR}.

Despite these attractive theoretical and practical features, accurate CVaR estimation remains computationally demanding.
Naive Monte Carlo (MC) sampling requires a prohibitively large number of forward model evaluations to adequately resolve tail
behavior, particularly as the confidence level approaches unity, rendering it impractical for high-fidelity finite
element simulations \cite{Hong2011MC}.
This challenge has motivated the development of classical rare-event simulation and variance-reduction techniques aimed
at concentrating computational effort in critical regions of the probability space.
Representative approaches include importance sampling \cite{Melchers1999Reliability}, subset simulation
\cite{Au2001SubsetSim,xie2023combined}, and surrogate-based response surface and polynomial chaos methods
\cite{owen2017comparison,Blatman2010AdaptivePC}.
While these methods can substantially improve efficiency relative to naive sampling, their performance may deteriorate
for complex mechanics problems involving high-dimensional correlated random fields, nonlinear constitutive behavior, and
computationally expensive finite element solvers \cite{Sudret2008GlobalSA,LeMa2010StochasticFEM}.
In particular, the generation of representative realizations of spatially correlated material fields using, for
example, Karhunen--Lo\`eve expansions \cite{Ghanem1991KL,Xiu2010UQBook} or low-rank Gaussian process and Nystr\"om kernel
approximations \cite{Williams2001Nystrom,Rasmussen2006GPML} requires careful numerical treatment to avoid prohibitive
memory and computational costs.
Consequently, even with advanced variance-reduction and surrogate modeling, the number of required forward simulations
can remain prohibitively large for high-confidence CVaR estimation.

Quantum amplitude estimation (QAE) has emerged as a foundational quantum algorithmic primitive for the evaluation of
bounded expectations and integrals, providing a provable asymptotic improvement in sampling complexity over classical
Monte Carlo under the standard oracle query model
\cite{Brassard2002QAE,Heinrich2002QuantumSummation,Montanaro2015MonteCarlo}.
In its seminal formulation, Brassard \textit{et al.} \cite{Brassard2002QAE} established amplitude amplification and
amplitude estimation as quantum analogues of classical averaging and numerical integration, demonstrating that
estimation error can scale inversely with the number of oracle queries rather than with its square root as in classical
sampling.
Subsequent complexity-theoretic analyses clarified the role of amplitude estimation as the quantum counterpart of Monte
Carlo summation and integration and established matching lower bounds for broad classes of expectation estimation
problems \cite{Heinrich2002QuantumSummation,Montanaro2015MonteCarlo}.
These results positioned amplitude estimation as a central building block for quantum-accelerated uncertainty
quantification and stochastic simulation.

The relevance of amplitude-estimation-based quantum algorithms to tail-sensitive risk measures has been explored
primarily in financial and probabilistic modeling, where value-at-risk and conditional value-at-risk can be expressed as
bounded hinge or loss expectations that are naturally compatible with amplitude estimation
\cite{Woerner2019QuantumFinance}.
In this context, Woerner and Egger \cite{Woerner2019QuantumFinance} formulated quantum algorithms for VaR and CVaR
computation and demonstrated the asymptotic sampling-complexity advantages of amplitude estimation for risk analysis.
Related quantum Monte Carlo approaches have further investigated expectation and tail-risk estimation for derivative
pricing and probabilistic simulation using amplitude-estimation primitives
\cite{Rebentrost2018PRA,Stamatopoulos2019OptionPricing,austin2012quantum}.
These developments establish amplitude estimation as a theoretically powerful computational engine for tail-risk
quantification.

Despite these asymptotic advantages, canonical amplitude estimation relies on quantum phase estimation and controlled
Grover amplification \cite{Grover1996Search,Brassard2002QAE}, resulting in deep circuits that are challenging to implement
on near-term quantum hardware.
This limitation has motivated the development of phase-estimation-free alternatives that trade modest classical
post-processing for substantial reductions in circuit depth and qubit overhead \cite{Suzuki2020QIP}.
Among these, iterative quantum amplitude estimation (IQAE) \cite{Grinko2021IQAE} has emerged as a particularly practical
approach, replacing phase estimation with adaptive confidence-interval refinement based on repeated Grover
amplification.
Subsequent extensions have introduced likelihood-based and Bayesian formulations to improve robustness under finite-shot
noise and device imperfections \cite{Suzuki2020QIP,Wiebe2016EfficientBayesianPE}.
Nevertheless, existing IQAE variants typically rely on feasibility-based interval updates and fixed or heuristic probing
schedules, which can exhibit instability in the presence of aliasing and finite-sample noise.

In parallel, there has been growing interest in applying quantum algorithms to problems in computational mechanics and
structural reliability \cite{he2025efficient, tabarraei2025variational, arora2025implementation}.
However, existing studies have largely focused on isolated algorithmic components or idealized demonstrations and have
not yet established quantum workflows that couple realistic material uncertainty models with full finite element
analyses.
In particular, quantum treatments of spatially correlated random fields and their propagation through high-fidelity
structural solvers remain largely unexplored.
While foundational connections between quantum algorithms and the finite element method have been established in a
theoretical setting \cite{Montanaro2016FEM}, and recent work has investigated quantum-enhanced pipelines for data-driven
mechanics, multiscale modeling, and topology optimization
\cite{Xu2024QCDataDriven,kim2025variational,tabarraei2025variational}, the integration of amplitude-estimation-based
inference with correlated material uncertainty and tail-risk objectives such as CVaR has not been systematically
addressed.
This gap motivates the present work.

In this work, we introduce a stabilized maximum-likelihood iterative quantum amplitude estimation (ML-IQAE) framework
for CVaR estimation of structural responses under correlated random fields.
ML-IQAE augments IQAE with likelihood-constrained interval tracking, adaptive Grover-depth and shot allocation,
multi-hypothesis pruning, periodic low-depth disambiguation, and explicit failure-probability budgeting, providing
statistically stable convergence under aliasing and finite-shot noise.
We adopt a finite-scenario framework in which a low-rank Nystr\"om approximation \cite{Williams2001Nystrom} is used to
generate spatially correlated material modulus fields for one-dimensional and two-dimensional benchmark problems.
Conditional on a fixed value-at-risk threshold determined from the scenario distribution, CVaR estimation reduces to the
expectation of a normalized hinge function bounded in $[0,1]$, which naturally admits an amplitude estimation
formulation \cite{Woerner2019QuantumFinance}.

We provide controlled comparisons against classical Monte Carlo under identical fixed VaR conditions and report absolute
error as a function of oracle-call budget.
All scenario responses are generated using full finite element solves under correlated random fields, and ML-IQAE
operates on a lookup-table oracle constructed from these high-fidelity simulations.
The results demonstrate that ML-IQAE exhibits sub-classical sampling complexity in practice, consistent with the
asymptotic convergence guarantees of amplitude estimation
\cite{Brassard2002QAE,Grinko2021IQAE}.
By integrating mechanics-relevant uncertainty models with a mechanically grounded CVaR formulation and stabilized
likelihood-constrained amplitude estimation, this work establishes a systematic benchmark at the intersection of
computational mechanics, structural reliability, and quantum algorithms, and provides a structured baseline for future
studies of quantum-accelerated tail-risk estimation in engineering.

\section{Problem formulation and reduction of CVaR estimation to amplitude estimation}

\subsection{Stochastic finite element formulation}

Let $\boldsymbol{x} \in \Omega \subset \mathbb{R}^d$ denote the spatial coordinate in the structural domain with boundary $\partial\Omega = \Gamma_u \cup \Gamma_t$, where $\Gamma_u$ and $\Gamma_t$ are the Dirichlet and Neumann boundaries, respectively. The stochastic linear elasticity problem seeks a displacement field $\mathbf{u}(\boldsymbol{x},\omega) \in \mathbb{R}^d$ satisfying
\begin{equation}
-\boldsymbol{\nabla} \cdot \left( \mathbf{C}(\boldsymbol{x},\omega) : \boldsymbol{\nabla}^s \mathbf{u}(\boldsymbol{x},\omega) \right) = \mathbf{f}(\boldsymbol{x}), \quad \boldsymbol{x} \in \Omega,
\end{equation}
with boundary conditions $\mathbf{u} = \mathbf{0}$ on $\Gamma_u$ and $\boldsymbol{\sigma}(\mathbf{u})\,\mathbf{n} = \mathbf{t}$ on $\Gamma_t$. Here $\mathbf{C}(\boldsymbol{x},\omega)$ denotes the random fourth-order elasticity tensor, $\boldsymbol{\nabla}^s(\cdot)$ is the symmetric gradient operator, $\boldsymbol{\sigma}$ is the Cauchy stress tensor, $\mathbf{n}$ is the outward unit normal on $\Gamma_t$, and $\mathbf{f}(\boldsymbol{x})$ and $\mathbf{t}(\boldsymbol{x})$ denote prescribed body forces and surface tractions, respectively. The random parameter $\omega$ is defined on the probability space $(\Xi,\mathcal{F},\mathbb{P})$, where $\Xi$ is the sample space, $\mathcal{F}$ is the associated $\sigma$-algebra, and $\mathbb{P}$ is the probability measure.

A conforming finite element discretization yields the stochastic linear system
\begin{equation}
\mathbf{K}(\omega)\,\mathbf{u}(\omega) = \mathbf{f},
\end{equation}
where $\mathbf{K}(\omega) \in \mathbb{R}^{N \times N}$ is the random global stiffness matrix assembled from element-level contributions, $\mathbf{u}(\omega) \in \mathbb{R}^{N}$ is the vector of nodal displacements, and $\mathbf{f} \in \mathbb{R}^{N}$ is the deterministic global load vector. For each realization, a scalar structural performance quantity of interest (QoI), such as the compliance, stress or displacement, is computed. 

\subsubsection{Nyström low-rank representation of correlated random fields}\label{nystrom}

The stochastic elasticity tensor $\mathbf{C}(\boldsymbol{x},\omega)$ is modeled through a spatially correlated
lognormal random field for the Young’s modulus $E(\boldsymbol{x},\omega)$. Specifically, we assume that the
log-modulus field admits the representation
\begin{equation}
\log E(\boldsymbol{x},\omega) = \sigma \sum_{j=1}^{r} \phi_j(\boldsymbol{x}) \, \xi_j(\omega),
\end{equation}
where $\{\xi_j\}_{j=1}^{r}$ are independent standard normal random variables and
$\{\phi_j(\boldsymbol{x})\}$ form a truncated low-rank spatial basis obtained from a Nyström approximation of a
Gaussian kernel covariance operator.

Let $k(\boldsymbol{x},\boldsymbol{x}')$ denote the anisotropic Gaussian covariance kernel
\begin{equation}
k(\boldsymbol{x},\boldsymbol{x}') =
\exp\!\left(
-\tfrac{1}{2}
\left[
\left(\tfrac{x_1-x_1'}{\ell_x}\right)^2 +
\left(\tfrac{x_2-x_2'}{\ell_y}\right)^2
\right]
\right).
\end{equation}
Given a set of spatial sampling points $\{\boldsymbol{x}_m\}_{m=1}^{M}$, the Nyström method constructs a low-rank
approximation of the kernel operator by eigen-decomposition of the matrix
$K_{MM}=[k(\boldsymbol{x}_m,\boldsymbol{x}_n)]$ and yields the approximate eigenfunctions
\begin{equation}
\phi_j(\boldsymbol{x}) \approx \sum_{m=1}^{M} k(\boldsymbol{x},\boldsymbol{x}_m)
\frac{u_{mj}}{\sqrt{\lambda_j}}, \qquad j=1,\dots,r,
\end{equation}
where $\lambda_j$ and $u_{mj}$ are eigenpairs of $K_{MM}$.

This representation enables efficient sampling of high-dimensional correlated random fields using only $r \ll N$
latent variables while preserving spatial correlation structure compatible with the finite element discretization.
Each realization $\omega_i$ therefore corresponds to a vector $\boldsymbol{\xi}^{(i)}\in\mathbb{R}^r$ which induces
a deterministic Young’s modulus field $E(\boldsymbol{x},\omega_i)$ and hence a deterministic stiffness matrix
$\mathbf{K}(\omega_i)$.

\subsection{Tail-risk measures}

Let $Q(\omega)$ denote a real-valued random structural response. The value-at-risk (VaR) at confidence level $\alpha \in (0,1)$ is defined as the lower $\alpha$-quantile of the response distribution \cite{rockafellar2000optimization, Pflug2000VaR}
\begin{equation}
\eta_\alpha = \inf \left\{ \eta \in \mathbb{R} \;:\; \mathbb{P}\!\left( Q(\omega) \le \eta \right) \ge \alpha \right\}.
\end{equation}
The conditional value-at-risk (CVaR), also referred to as the expected shortfall, measures the expected severity of responses exceeding the VaR threshold and is defined as \cite{Acerbi2002CVaR,rockafellar2002conditional}
\begin{equation}
\mathrm{CVaR}_\alpha(Q) = \mathbb{E}\!\left[ Q(\omega) \,\middle|\, Q(\omega) \ge \eta_\alpha \right].
\end{equation}
An equivalent convex optimization-based characterization of CVaR is given by \cite{rockafellar2002conditional,rockafellar2000optimization}
\begin{equation}
\mathrm{CVaR}_\alpha(Q)
= \min_{\eta \in \mathbb{R}}
\left\{
\eta + \frac{1}{1-\alpha} \, \mathbb{E}\!\left[ (Q(\omega)-\eta)_+ \right]
\right\},
\end{equation}
where $(x)_+ = \max(x,0)$ denotes the positive part operator.

\subsection{Discrete scenario formulation}
To enable computational treatment, the continuous probability space is approximated by a finite discrete scenario set
\begin{equation}
\{(\omega_i,p_i)\}_{i=1}^{N_s}, \qquad p_i > 0, \quad \sum_{i=1}^{N_s} p_i = 1,
\end{equation}
where $\omega_i$ denotes a realization of the underlying random field and $p_i$ is the associated probability weight. The corresponding structural responses are denoted by
\begin{equation}
Q_i = Q(\omega_i), \qquad i = 1,\ldots,N_s .
\end{equation}

Let $\{Q_{(i)}\}$ denote the ordered response values such that
\begin{equation}
Q_{(1)} \le Q_{(2)} \le \cdots \le Q_{(N_s)},
\end{equation}
with reordered probabilities $\{p_{(i)}\}$. The discrete value-at-risk at confidence level $\alpha$ is then defined as the smallest response level satisfying
\begin{equation}
\sum_{j=1}^{k_\alpha} p_{(j)} \ge \alpha,
\end{equation}
and is given by
\begin{equation}
\eta_\alpha = Q_{(k_\alpha)},
\end{equation}
where
\begin{equation}
k_\alpha = \min \left\{ k : \sum_{j=1}^{k} p_{(j)} \ge \alpha \right\}.
\end{equation}

Using this discrete VaR threshold, the conditional value-at-risk is expressed as
\begin{equation}
\mathrm{CVaR}_\alpha(Q) = \eta_\alpha + \frac{1}{1-\alpha} \sum_{i=1}^{N_s} p_i \max(Q_i - \eta_\alpha, 0),
\label{eq:discrete_cvar}
\end{equation}
which is a discrete realization of the optimization-based CVaR representation.

\subsection{Reduction of CVaR to a bounded expectation}

Equation \eqref{eq:discrete_cvar} shows that CVaR estimation reduces to computing the expectation of a hinge function of the structural response. Quantum amplitude estimation is naturally suited to estimating expectations of bounded random variables. To bring the discrete CVaR formulation into a form compatible with amplitude estimation, following the normalization strategy introduced in quantum risk analysis \cite{Woerner2019QuantumFinance,Rebentrost2018PRA}, we define normalized hinge variables
\begin{equation}
g_i = \frac{\max(Q_i - \eta_\alpha,0)}{Q_{\max}-\eta_\alpha}, 
\qquad Q_{\max} = \max_i Q_i,
\end{equation}
which satisfy $g_i \in [0,1]$.The normalized expectation
\begin{equation}
a = \sum_{i=1}^{N_s} p_i g_i = \mathbb{E}[g(\omega)]
\label{aExpectation}
\end{equation}
can be directly encoded as a quantum amplitude and estimated via amplitude estimation.

The original CVaR is then recovered by the affine transformation
\begin{equation}
\mathrm{CVaR}_\alpha(Q)
= \eta_\alpha + \frac{Q_{\max}-\eta_\alpha}{1-\alpha} \, a,
\end{equation}
so that CVaR estimation reduces to the quantum estimation of a bounded expectation $a \in [0,1]$.

\section{Quantum algorithm for CVaR estimation}

It was shown in \eref{aExpectation} that CVaR estimation for the stochastic finite element model reduces to the computation of a single bounded expectation $a \in [0,1]$. This quantity represents the normalized expected contribution of tail responses beyond the value-at-risk threshold and fully determines the desired conditional value-at-risk through an affine transformation. Consequently, the sole objective of the quantum algorithm developed in this work is to estimate the scalar parameter $a$ with high accuracy using as few effective evaluations of the underlying stochastic model as possible.

Iterative quantum amplitude estimation (IQAE) provides a principled quantum inference framework for this task. Classical Monte Carlo estimates $a$ by repeatedly sampling realizations of $g(\omega)$ and forming an empirical average, which leads to an estimation error that decays only at a rate proportional to the inverse square root of the number of samples. In contrast, quantum amplitude estimation exploits coherent quantum superposition and interference to encode the entire discrete probability distribution into a single quantum state, enabling estimation of bounded expectations with a mean-square error that scales inversely with the number of oracle queries under the standard quantum query model \cite{Brassard2002QAE,Montanaro2015MonteCarlo}. This quadratic improvement in sampling complexity positions amplitude estimation as the quantum analogue of classical Monte Carlo integration and summation \cite{Heinrich2002QuantumSummation}.

IQAE is a hardware-efficient reformulation of canonical amplitude estimation that avoids quantum phase estimation and deep controlled-unitary constructions, replacing them with an adaptive classical interval inference procedure driven by repeated Grover-type amplifications and projective measurements \cite{Grinko2021IQAE}. In the present context, IQAE operates by first embedding the discrete normalized tail-response values $\{g_i\}$ into a quantum state via a state-preparation oracle that coherently represents all stochastic scenarios and their associated probabilities. A Grover amplification operator is then applied iteratively to rotate the quantum state within a low-dimensional invariant subspace, systematically amplifying the probability amplitude associated with the desired expectation parameter $a$. By performing measurements at different amplification depths and combining the resulting statistics using rigorous confidence interval constructions, IQAE progressively refines a feasible set for $a$ while maintaining robustness to finite-shot noise and measurement uncertainty.

The structure and physical interpretation of the oracle construction and the Grover amplification operator are detailed in the following subsections.

\subsection{Oracle construction}
\label{sec:oracle_construction}

The role of the oracle in quantum amplitude estimation is to prepare a quantum state in which a target bounded expectation appears explicitly as the probability of a measurable binary event. In contrast to Boolean decision oracles, the oracle employed here is a \emph{state-preparation oracle} that encodes real-valued information into quantum probability amplitudes.

In the present setting, the objective is to estimate the scalar expectation
\begin{equation}
a = \sum_{i=1}^{N_s} p_i g_i ,
\label{eq:a_def}
\end{equation}
where $\{g_i\}_{i=1}^{N_s}$ are normalized tail-response values and $\{p_i\}_{i=1}^{N_s}$ are the corresponding scenario probabilities.

Let $N_s$ denote the number of discrete stochastic scenarios and let
\begin{equation}
n = \lceil \log_2 N_s \rceil
\end{equation}
be the number of qubits required to index them. The oracle is implemented as a unitary operator $A$ acting on $n$ index qubits and one ancilla qubit, constructed such that
\begin{equation}
A\ket{0}
=
\sum_{i=1}^{N_s} \sqrt{p_i}\ket{i}
\Big(
\sqrt{1-g_i}\ket{0}
+
\sqrt{g_i}\ket{1}
\Big).
\label{eq:oracle_def}
\end{equation}

Measuring the ancilla qubit yields outcome $\ket{1}$ with probability
\begin{equation}
\Pr(\text{ancilla}=1)=a ,
\label{eq:a_prob}
\end{equation}
so that estimation of the expectation $a$ reduces to inference of a single quantum measurement probability.

Define the projectors
\begin{equation}
\Pi_1 = I \otimes \ket{1}\!\bra{1}, \qquad
\Pi_0 = I \otimes \ket{0}\!\bra{0}.
\end{equation}
Then
\begin{equation}
a = \bra{0} A^\dagger \Pi_1 A \ket{0},
\end{equation}
and the oracle-prepared state $\ket{\Psi} = A\ket{0}$ admits the orthogonal decomposition
\begin{equation}
\ket{\Psi}
=
\sqrt{1-a}\,\ket{\psi_0}
+
\sqrt{a}\,\ket{\psi_1},
\label{eq:psi_decomp}
\end{equation}
where
\begin{equation}
\ket{\psi_1} = \frac{\Pi_1 \ket{\Psi}}{\sqrt{a}}, \qquad
\ket{\psi_0} = \frac{\Pi_0 \ket{\Psi}}{\sqrt{1-a}}.
\end{equation}

Introducing a global angle $\theta \in (0,\pi/2)$ such that
\begin{equation}
\sin^2 \theta = a = \sum_{i=1}^{N_s} p_i g_i,
\label{eq:theta_def}
\end{equation}
the oracle action can be written compactly as
\begin{equation}
A\ket{0}
=
\cos \theta\,\ket{\psi_0}
+
\sin \theta\,\ket{\psi_1}.
\label{eq:oracle_theta_form}
\end{equation}

Equation~\eqref{eq:oracle_theta_form} defines the two-dimensional invariant subspace
\begin{equation}
\mathcal{K} = \mathrm{span}\{\ket{\psi_0}, \ket{\psi_1}\},
\label{eq:K_def}
\end{equation}
within which all Grover amplification dynamics take place.
\begin{figure*}[tb]
  \centering
  \includegraphics[width=0.7\linewidth]{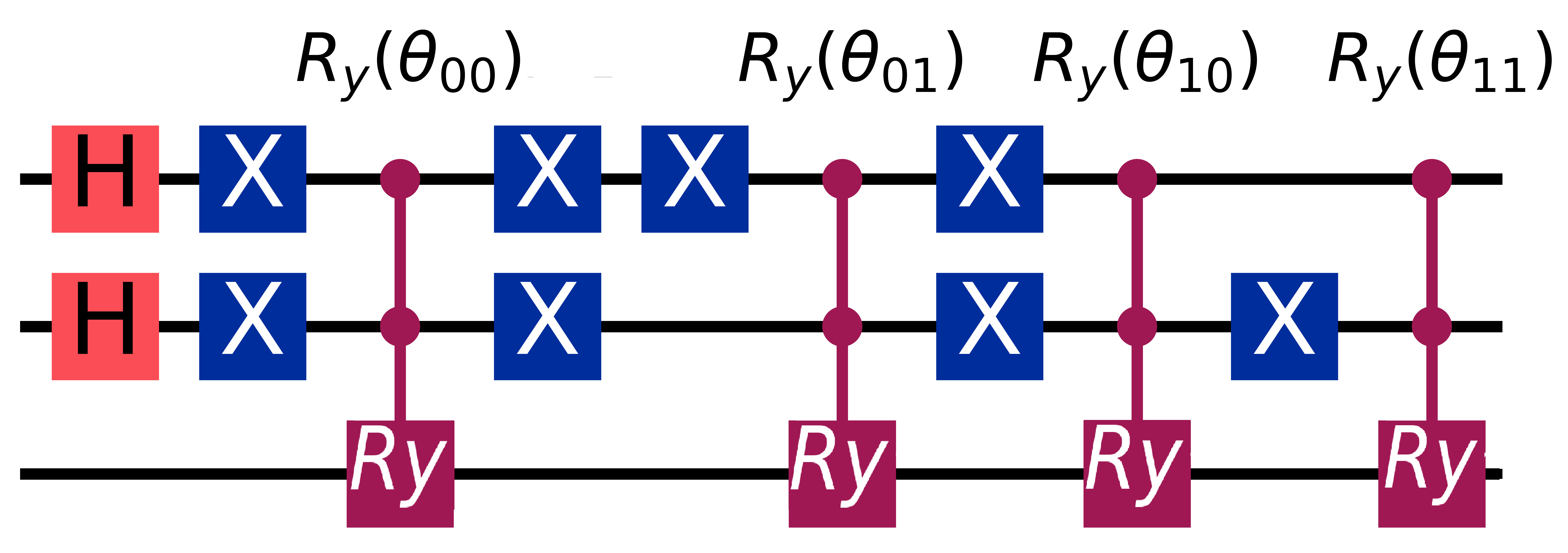}
  \caption{Oracle-$A$ circuit used in this work to encode scenario-dependent values $g_i$ into ancilla amplitudes via scenario-controlled $R_y$ rotations. Pattern conversions required for controls-on-$|0\rangle$ are implemented using surrounding $X$ gates, yielding a compact ``lookup-table'' implementation.}
  \label{fig:oracleA_circuit}
\end{figure*}
\subsubsection{Circuit-level realization and gate-level encoding}
\label{sec:oracle_circuit}

Figure~\ref{fig:oracleA_circuit} illustrates the circuit realization of the oracle $A$. In general, the oracle acts on an $n$-qubit index register together with a single ancilla qubit. For clarity and concreteness, Fig.~\ref{fig:oracleA_circuit} depicts the construction for a two-qubit index register, corresponding to four discrete scenarios, with the ancilla implemented as qubit~2. The index qubits encode the scenario label, while the ancilla qubit is used to store the amplitude information required for quantum amplitude estimation.

The oracle circuit implements a conditional amplitude-encoding mechanism in which the normalized scenario-dependent values $\{g_i\}_{i=1}^{N_s}$ are mapped to the probability of measuring the ancilla qubit in state $\ket{1}$. This is achieved by applying scenario-controlled single-qubit rotations to the ancilla qubit, conditioned on the computational basis state of the index register.

For each scenario $i$, the oracle applies a controlled rotation $R_y(\phi_i)$ to the ancilla qubit, where the $R_y$ gate is defined by the matrix
\begin{equation}
R_y(\phi_i)
=
\begin{pmatrix}
\cos(\phi_i/2) & -\sin(\phi_i/2) \\
\sin(\phi_i/2) & \cos(\phi_i/2)
\end{pmatrix}.
\label{eq:Ry_matrix}
\end{equation}
When acting on the ancilla state $\ket{0}$, this rotation produces the superposition
\begin{equation}
R_y(\phi_i)\ket{0}
=
\cos(\phi_i/2)\ket{0}
+
\sin(\phi_i/2)\ket{1}.
\end{equation}
By the Born measurement rule, measuring the ancilla qubit then yields outcome $\ket{1}$ with probability
\begin{equation}
\Pr(\text{ancilla}=1 \mid i) = \sin^2(\phi_i/2).
\end{equation}
Accordingly, the rotation angles are chosen such that
\begin{equation}
\sin^2(\phi_i/2) = g_i,
\label{eq:phi_def}
\end{equation}
ensuring that the ancilla measurement probability reproduces the prescribed normalized response value for scenario $i$.

The rotations $R_y(\phi_i)$ are implemented as multi-controlled operations in which the index qubits serve as controls and the ancilla qubit is the target. Each rotation is designed to act only when the index register is in a specific computational basis state corresponding to scenario $i$. Since standard controlled gates activate on control qubits in the $\ket{1}$ state, controls on $\ket{0}$ are realized by temporarily applying Pauli-$X$ gates to the relevant index qubits before and after the controlled rotation. This control-pattern conversion ensures that exactly one basis state of the index register activates each ancilla rotation, while the index register is restored to its original configuration after the operation.

Taken together, the circuit realizes a coherent lookup-table oracle; each basis state of the index register activates a specific ancilla rotation that encodes the associated value $g_i$. When the index register is prepared in a superposition, all such controlled rotations act coherently, producing the oracle state in Eq.~\eqref{eq:oracle_def}. We emphasize that the rotation angles $\{\phi_i\}$ are gate-level parameters used solely for local amplitude encoding within the circuit and are distinct from the global amplitude-estimation angle $\theta$ defined by $\sin^2\theta = a$, which governs the subsequent Grover amplification dynamics.

\begin{figure}[t]
  \centering
  \includegraphics[width=0.6\linewidth]{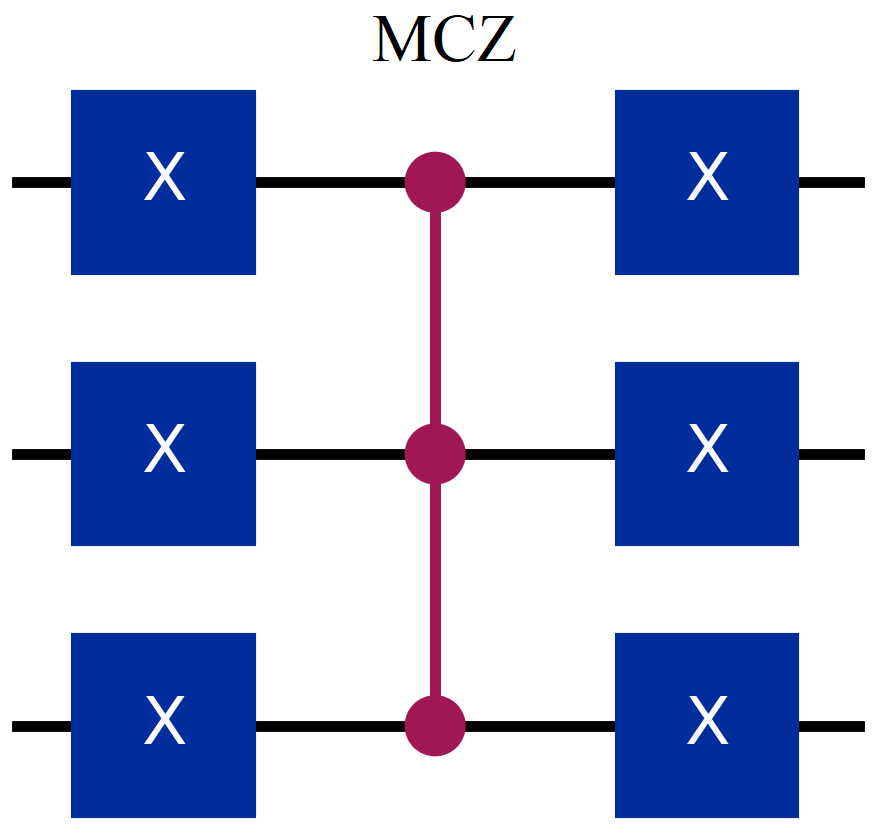}
\caption{Circuit implementation of the reflection 
$S_0 = I - 2|000\rangle\langle 000|$ using Pauli-$X$ conjugation of a
multi-controlled $Z$ gate.}
  \label{fig:S0_circuit}
\end{figure}

\begin{figure*}[t]
  \centering
  \includegraphics[width=\linewidth]{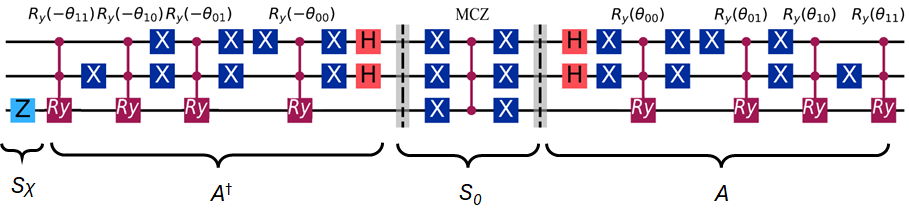}
 \caption{Quantum circuit for one Grover amplification step
$G = -A S_0 A^\dagger S_\chi$.
The circuit consists of the success reflection $S_\chi$, the inverse oracle
$A^\dagger$, the zero-state reflection $S_0$ implemented with a
multi-controlled $Z$ gate, and the oracle $A$.}
  \label{fig:grover_circuit}
\end{figure*}

\subsection{Grover amplification}
\label{sec:grover}

Grover amplification \cite{Grover1996Search, Brassard2002QAE, NielsenChuang2010, Montanaro2015MonteCarlo} exploits the two-dimensional invariant subspace $\mathcal{K}$
defined in Eq.~\eqref{eq:K_def} to perform deterministic and reversible rotations
that amplify information associated with the unknown amplitude-estimation angle
$\theta$. When restricted to this subspace, the algorithm admits a geometric
interpretation in terms of successive reflections that coherently rotate the
oracle-prepared state toward the success subspace associated with the ancilla
measurement outcome.

The two reflection operators underlying Grover amplification are
\begin{equation}
S_\chi = I - 2\ket{\psi_1}\!\bra{\psi_1},
\qquad
S_0 = I - 2\ket{0}\!\bra{0},
\label{eq:reflections}
\end{equation}
where $S_\chi$ implements a reflection about the success subspace spanned by
$\ket{\psi_1}$ and $S_0$ implements a reflection about the all-zero computational
basis state. In the present setting, success is defined by observing the ancilla
qubit in state $\ket{1}$, and $S_\chi$ therefore acts as a selective phase flip on
all basis states with ancilla bit equal to one. At the circuit level, this
operation is realized by a single Pauli-$Z$ gate acting on the ancilla qubit,
which introduces a $\pi$ phase shift on all computational basis states with
ancilla value $|1\rangle$.

The reflection $S_0$ is realized as a multi-controlled phase flip that applies a
$\pi$ phase shift exclusively to the computational basis state
$\ket{0}^{\otimes(n+1)}$, where the additional qubit corresponds to the ancilla.
As illustrated in Fig.~\ref{fig:S0_circuit}, this operation is implemented by first
applying Pauli-$X$ gates to all qubits to map $\ket{0}$ to
$\ket{1}^{\otimes(n+1)}$, followed by a multi-controlled $Z$ (MCZ) gate, and finally
uncomputing the basis transformation with a second layer of Pauli-$X$ gates. The
resulting action satisfies
\[
\ket{0} \mapsto -\ket{0}, \qquad
\ket{x} \mapsto \ket{x}, \quad x \neq 0,
\]
which is precisely the defining action of $S_0$.

Combining the two reflections with the oracle $A$, the Grover iterate is defined as
\begin{equation}
G = -A S_0 A^\dagger S_\chi .
\label{eq:grover_op}
\end{equation}
A circuit-level realization of this operator, including the implementations of
both $S_\chi$ and $S_0$, is shown in Fig.~\ref{fig:grover_circuit}.

Restricted to the invariant subspace $\mathcal{K}$, the Grover operator acts as a
planar rotation by angle $2\theta$. After $k$ Grover iterations, the state evolves
according to
\begin{equation}
G^k A\ket{0}
=
\cos\!\big((2k+1)\theta\big)\ket{\psi_0}
+
\sin\!\big((2k+1)\theta\big)\ket{\psi_1},
\label{eq:grover_k}
\end{equation}
where $\ket{\psi_0}$ and $\ket{\psi_1}$ denote the orthogonal components of the
oracle-prepared state within $\mathcal{K}$. Consequently, the probability of
observing the ancilla qubit in the success state after $k$ Grover iterations is
\begin{equation}
p_k(\theta)
=
\sin^2\!\big((2k+1)\theta\big).
\label{eq:pk_def}
\end{equation}

Equations~\eqref{eq:grover_k} and \eqref{eq:pk_def} show that Grover amplification
transforms estimation of the static expectation value $a=\sin^2\theta$ into
estimation of the oscillation frequency $\theta$. Interval-based quantum amplitude
estimation leverages this deterministic oscillatory structure to infer $a$ from
amplified measurement statistics while maintaining rigorous finite-sample
confidence guarantees.






\section{Iterative quantum amplitude estimation for CVaR inference}

Rather than estimating the unknown rotation angle $\theta$ via quantum phase estimation,
which requires deep controlled-unitary circuits and additional ancilla qubits and forms
the basis of the original amplitude estimation algorithm introduced by Brassard~et~al.\cite{Brassard2002QAE}, we employ iterative quantum amplitude estimation (IQAE) \cite{Grinko2021IQAE, Suzuki2020QIP}.
IQAE avoids the quantum Fourier transform and phase estimation entirely, instead using
carefully selected Grover amplification and classical post-processing to infer the
amplitude from repeated measurements with rigorous confidence bounds.
This approach significantly reduces the depth and qubit requirements compared to
phase-estimation-based methods while retaining near-optimal query complexity.

At amplification depth $k$, the amplified quantum state
\begin{equation}
\ket{\psi_k} = G^k A \ket{0}
\end{equation}
is prepared and measured multiple times. The probability of observing the ancilla in the success state is given by \eref{eq:pk_def}. By collecting measurement statistics across multiple amplification depths, IQAE progressively restricts the set of admissible parameter values that are statistically consistent with the observed data.

Throughout the IQAE procedure, we prescribe a global failure tolerance
$\delta_{\mathrm{tot}}\in(0,1)$ that bounds the probability that the maintained admissible parameter set ever excludes the true rotation angle $\theta$ due to finite-shot statistical fluctuations. All per-round confidence allocations and disambiguation steps introduced below are constructed such that the overall probability of incorrectly excluding the true parameter satisfies
\[
\Pr\!\left(\theta \notin \Theta^{(t)} \text{ for some } t\right) \le \delta_{\mathrm{tot}}.
\]
This global tolerance serves as the primary confidence parameter governing the statistical reliability of the entire inference procedure and underpins the confidence--constrained estimation framework developed in the subsequent sections.

\subsection{Measurement model and likelihood}
\label{sec:iqae_measurement_model}

Grover amplification maps the bounded expectation $a=\sin^2\theta$ to the oscillatory success probability $p_k(\theta)$ given in \eref{eq:pk_def}. Estimating CVaR therefore, reduces to statistical inference of the unknown rotation angle $\theta$ from ancilla measurements of amplified quantum states.

At IQAE round $t$, an amplification depth $k_t$ is selected and the amplified state $G^{k_t}A\ket{0}$ is executed $m_t$ times. Each execution produces a Bernoulli outcome
\[
Y_{t,i}\sim \mathrm{Bernoulli}(p_{k_t}(\theta)), \qquad i=1,\ldots,m_t,
\]
where $Y_{t,i}=1$ corresponds to observing the ancilla in the success state $\ket{1}$. Let $h_t=\sum_{i=1}^{m_t} Y_{t,i}$ denote the number of observed successes at round $t$. Then
\[
h_t \sim \mathrm{Binomial}(m_t,p_{k_t}(\theta)), \qquad \hat p_{k_t}=\frac{h_t}{m_t}.
\]

Let
\begin{equation}
\mathcal{D}_t=\{(k_j,m_j,h_j)\}_{j=1}^{t}
\label{eq:Dt_def}
\end{equation}
denote the measurement data collected up to and including round $t$. Assuming independent state preparations and measurements across rounds, the likelihood of the unknown parameter $\theta$ given $\mathcal{D}_t$ is
\begin{equation}
\mathcal{L}(\theta;\mathcal{D}_t)
=
\prod_{j=1}^{t}
\binom{m_j}{h_j}
\big[p_{k_j}(\theta)\big]^{h_j}
\big[1-p_{k_j}(\theta)\big]^{m_j-h_j}.
\label{eq:joint_likelihood}
\end{equation}

Equation \eqref{eq:joint_likelihood} casts IQAE as a nonlinear binomial regression problem in which $\theta$ is inferred from noisy observations of the response family $\{p_k(\theta)\}_{k\ge0}$.

The online likelihood model \eqref{eq:joint_likelihood} is operationalized by the adaptive interval-based procedure summarized in Algorithm~\ref{alg:iqae_round}. The resulting interval updates, hypothesis management and periodic disambiguation are developed in the subsequent subsections.

\subsubsection{Maximum-likelihood estimation within admissible hypotheses}
\label{sec:mle_iqae}

The interval-based inference procedure described above maintains a union of admissible parameter intervals
$\Theta^{(t)} \subset (0,\pi/2)$ that contains the true rotation angle $\theta$ with prescribed global confidence.
While this feasible set provides rigorous uncertainty certification, point estimation of $\theta$ within this set
is required to obtain a sharp estimator for the underlying amplitude $a=\sin^2\theta$.

\begin{algorithm}[tb]
\caption{Structure of one ML-IQAE round $t$ (feasibility update and MLE point estimate)}
\label{alg:iqae_round}
\begin{algorithmic}[1]
\State Select Grover depth $k_t$ based on the current admissible set $\Theta^{(t-1)}$.
\State Select number of shots $m_t$ subject to remaining oracle-call budget.
\State Execute the amplified circuit $m_t$ times at depth $k_t$.
\State Record the number of observed successes $h_t$.
\State Allocate local confidence budget $\delta_t$ via \eref{eq:delta_schedule}.
\State Construct a Clopper--Pearson confidence interval for $p_{k_t}(\theta)$.
\State Update admissible parameter sets $\Theta^{(t)}$ via interval intersection (with pruning to $M_{\max}$ components).
\State Apply disambiguation or restart logic if $\Theta^{(t)}=\emptyset$.
\State Update point estimate: compute $\hat\theta_t=\arg\max_{\theta\in\Theta^{(t)}} \ell(\theta;D_t)$ and $\hat a_t=\sin^2\hat\theta_t$.
\end{algorithmic}
\end{algorithm}

Given the accumulated measurement data $D_t$, 
we define the log-likelihood function induced by the binomial observation model \eref{eq:joint_likelihood} as
\begin{equation}
\ell(\theta;D_t)
=
\sum_{j=1}^{t}
\Big[
h_j \log p_{k_j}(\theta)
+
(m_j-h_j)\log\!\big(1-p_{k_j}(\theta)\big)
\Big],
\label{eq:mle_loglik}
\end{equation}
where $p_k(\theta)=\sin^2((2k+1)\theta)$.

The reported estimate of the rotation angle is defined as the constrained maximum-likelihood estimator
\begin{equation}
\hat\theta_t
=
\arg\max_{\theta\in\Theta^{(t)}} \ell(\theta;D_t),
\label{eq:mle_theta}
\end{equation}
with the corresponding amplitude estimate
\begin{equation}
\hat a_t=\sin^2\hat\theta_t.
\label{eq:mle_a}
\end{equation}

The admissible set $\Theta^{(t)}$ acts as a statistically valid feasibility constraint, ensuring that the true parameter
$\theta$ is excluded only with probability at most $\delta_{\mathrm{tot}}$, while the maximum-likelihood criterion
selects the most probable parameter value consistent with the accumulated data. This hybrid confidence--constrained
formulation combines the global correctness guarantees of interval-based IQAE with the asymptotic efficiency and
low-variance properties of maximum-likelihood estimation.

Consequently, the proposed estimator constitutes a confidence-constrained maximum-likelihood amplitude estimator,
rather than a purely feasibility-based IQAE interval midpoint rule. This distinction is fundamental to the improved
convergence behavior and variance reduction observed in the numerical experiments reported in Section~5.

\subsection{Interval-based feasibility tracking with hypothesis management and disambiguation}
\label{sec:iqae_unified}

The proposed ML-IQAE framework employs an interval-based feasibility-tracking procedure that progressively constructs
a confidence-consistent admissible region for the unknown rotation angle $\theta$. At each round, a new batch of
amplified measurements is acquired at a selected Grover depth, converted into a binomial confidence interval for the
corresponding success probability, and mapped into an admissible set of $\theta$ values. These admissible sets are
intersected with previously retained constraints to form a global feasible region that serves as a statistically valid
confidence certificate for the true parameter value.

Due to the oscillatory and non-injective structure of the response function
$p_k(\theta)=\sin^2((2k+1)\theta)$, the feasible region may consist of multiple disconnected components that are
simultaneously consistent with the accumulated data. ML-IQAE therefore maintains a union of disjoint admissible
intervals, referred to as hypotheses, and updates them sequentially as new measurements arrive. Spurious hypotheses are
periodically pruned using low-depth disambiguation measurements, while rare empty-set events are handled through the
controlled restart mechanism.

Since ML-IQAE proceeds sequentially under a finite quantum computational budget, we quantify cost directly in the query
model by explicitly accounting for oracle invocations. At round $t$, the algorithm selects an amplification depth $k_t$
and performs $m_t$ shots of the amplified circuit $G^{k_t}A\ket{0}$. Each shot applies the Grover iterate $G$ exactly
$k_t$ times and includes one initial application of $A$, so the oracle-call cost of round $t$ is
\begin{equation}
C_t=(2k_t+1)m_t.
\label{eq:oracle_cost_round}
\end{equation}
If the run terminates after $T$ rounds, the total oracle usage satisfies
\begin{equation}
\sum_{t=1}^{T} C_t \le B .
\label{eq:oracle_budget}
\end{equation}

Algorithm~\ref{alg:iqae_round} summarizes the feasibility-update procedure with hypothesis management that provides the
confidence-constrained admissible region used by the maximum-likelihood estimator defined in
Section~\ref{sec:mle_iqae}. The following subsections formalize its main components: (i) binomial confidence-interval
inference that maps each measurement batch into admissible sets for $\theta$, (ii) multi-hypothesis tracking to
represent disconnected feasible regions, and (iii) periodic low-depth disambiguation for pruning spurious modes.

All confidence intervals used by ML-IQAE are constructed under a global summable risk allocation
$\{\delta_t\}_{t\ge1}$ satisfying $\sum_{t\ge1}\delta_t\le\delta_{\mathrm{tot}}$,
which guarantees global correctness under adaptive measurement and hypothesis pruning.
The formal construction is given in Section~\ref{sec:iqae_core}.

\subsubsection{Binomial confidence bounds for amplified measurements}
\label{sec:binomial_ci}

At each round $t$, ML-IQAE converts the observed success count $h_t$ from $m_t$ amplified
measurements into a statistically valid confidence interval for the corresponding success
probability $p_{k_t}(\theta)$.
To ensure exact, non-asymptotic coverage under finite sampling and adaptive measurement
selection, we employ the classical Clopper--Pearson construction
\cite{Clopper1934Confidence,Brown2001Interval}.

Given a prescribed per-round failure tolerance $\delta_t$, the Clopper--Pearson confidence
interval $[p_L^{(t)},p_U^{(t)}]$ is defined by
\begin{equation}
p_L^{(t)} = I^{-1}_{\delta_t/2}\!\big(h_t,\; m_t-h_t+1\big),
\qquad
p_U^{(t)} = I^{-1}_{1-\delta_t/2}\!\big(h_t+1,\; m_t-h_t\big),
\label{eq:CP_def_main}
\end{equation}
Here $I^{-1}_q(a,b)$ denotes the inverse of the regularized incomplete beta function
\cite{AbramowitzStegun,NISTDLMF}, which arises naturally from inversion of the binomial
distribution in the Clopper--Pearson construction \cite{Brown2001Interval}.
This interval satisfies the exact coverage guarantee
\begin{equation}
\Pr\!\big(p_{k_t}(\theta)\in[p_L^{(t)},p_U^{(t)}]\big)\ge 1-\delta_t,
\label{eq:CP_coverage}
\end{equation}
independently of the sample size $m_t$ and without asymptotic assumptions.

The resulting probability interval $[p_L^{(t)},p_U^{(t)}]$ is subsequently mapped into an
admissible parameter set $\Theta_{k_t}$ through the response relation
$p_{k_t}(\theta)=\sin^2((2k_t+1)\theta)$, as formalized in
Section~\ref{sec:iqae_core}.

\subsubsection{Core interval inference}
\label{sec:iqae_core}
Let $\delta_{\mathrm{tot}}\in(0,1)$ denote the prescribed overall failure tolerance for a single ML-IQAE run.
We allocate a per-round risk budget $\delta_t$ such that
\begin{equation}
\sum_{t=1}^{\infty}\delta_t \le \delta_{\mathrm{tot}}.
\label{eq:delta_budget}
\end{equation}
In this work we employ the summable schedule
\begin{equation}
\delta_t=\frac{6}{\pi^2}\,\frac{\delta_{\mathrm{tot}}}{t^2},
\label{eq:delta_schedule}
\end{equation}
which satisfies \eqref{eq:delta_budget}.
All binomial confidence intervals used by ML-IQAE are constructed with per-round coverage at least $1-\delta_t$.

For the batch $(k_t,m_t,h_t)\in\mathcal{D}_t$, the empirical success frequency $\hat p_{k_t}=h_t/m_t$ estimates the response
probability $p_{k_t}(\theta)$ in \eqref{eq:pk_def}. A confidence interval $[p_L,p_U]$ is constructed using the exact
Clopper--Pearson method \cite{Clopper1934Confidence,Brown2001Interval} such that
\begin{equation}
\Pr\!\big(p_{k_t}(\theta)\in[p_L,p_U]\big)\ge 1-\delta_{t}.
\label{eq:pk_ci}
\end{equation}
where $[p_L,p_U]$ is computed from $(m_t,h_t)$. The induced admissible parameter set is
\begin{equation}
\Theta_{k_t}
=\Big\{\theta\in(0,\pi/2):p_{k_t}(\theta)\in[p_L,p_U]\Big\},
\label{eq:theta_set}
\end{equation}
which serves as a confidence-consistent feasibility constraint for the maximum-likelihood estimator defined in
Section~\ref{sec:mle_iqae}.

After $t$ rounds, the globally admissible region is
\begin{equation}
\Theta^{(t)}=\bigcap_{j=1}^{t} \Theta_{k_j},
\label{eq:theta_intersection}
\end{equation}
which constitutes a confidence-certified feasible region for $\theta$. For reporting uncertainty bounds, we also
maintain the convex hull
\begin{equation}
[\theta_L^{(t)},\theta_U^{(t)}]=\mathrm{hull}\ \!\big(\Theta^{(t)}\big).
\label{eq:theta_hull}
\end{equation}
By a union bound, the maintained hull interval satisfies
\begin{equation}
\Pr\!\left(\theta\in[\theta_L^{(t)},\theta_U^{(t)}]\right)
\ge 1-\sum_{j=1}^{t}\delta_{k_j}.
\label{eq:theta_coverage}
\end{equation}

\subsubsection{Adaptive selection of Grover depth}
\label{sec:adaptive_k}

At each round $t$, the Grover amplification depth $k_t$ is chosen adaptively based on the current admissible region
$\Theta^{(t-1)}$ and the accumulated measurement data. A provisional point estimate
$\hat\theta_{t-1}$ is first obtained by maximum-likelihood inference over $\Theta^{(t-1)}$, and a nominal amplification
depth is computed using the standard information-optimal probing condition of quantum amplitude estimation \cite{Brassard2002QAE,Suzuki2020QIP,Grinko2021IQAE}
\begin{equation}
k_{\mathrm{prop}}
=\left\lfloor \frac{\pi}{8\hat\theta_{t-1}}-\frac{1}{2} \right\rfloor .
\label{eq:k_prop}
\end{equation}
To prevent aliasing and loss of statistical identifiability due to the periodic and non-injective structure of
$p_k(\theta)=\sin^2((2k+1)\theta)$, we impose the alias-safety condition
\begin{equation}
(2k_t+1)\,\theta_U^{(t-1)} \le \kappa ,
\qquad \kappa = 0.49\pi ,
\label{eq:alias_safety}
\end{equation}
which guarantees that $p_k(\theta)$ remains strictly monotone over the entire admissible region. This yields the
corresponding safety bound
\begin{equation}
k_{\mathrm{safe}}
=\left\lfloor \frac{\kappa}{2\,\theta_U^{(t-1)}}-\frac{1}{2} \right\rfloor .
\label{eq:k_safe}
\end{equation}
The final depth is then selected as
\begin{equation}
k_t = \min\{k_{\mathrm{prop}},\,k_{\mathrm{safe}},\,k_{\max},\,k_{t-1}+1\},
\label{eq:k_select}
\end{equation}
where $k_{\max}$ is a prescribed upper bound and the monotonicity constraint $k_t\le k_{t-1}+1$ prevents abrupt jumps
that could destabilize the inference. In addition, if successive empirical success frequencies approach saturation near
$0$ or $1$, the algorithm temporarily reduces $k_t$ to avoid degenerate binomial observations. This adaptive policy
concentrates oracle calls at statistically informative amplification depths while preserving global robustness.

At each round $t$, the number of measurement shots $m_t$ is selected adaptively to balance statistical accuracy with
overall oracle-call efficiency. Because the binomial confidence interval width scales as
$O(m_t^{-1/2})$, increasing $m_t$ improves the precision of the induced admissible region $\Theta_t$, but excessive
shot counts at early rounds are inefficient when the feasible region is still wide.

ML-IQAE therefore employs a gradually increasing shot policy in which $m_t$ is chosen as a smooth function of the
remaining oracle-call budget and the current round index. This strategy allocates relatively few shots at early rounds
to obtain coarse but informative constraints, and progressively increases $m_t$ as the admissible region contracts and
higher resolution is required. The shot count is additionally capped by the remaining budget and a prescribed minimum
to ensure that each batch provides a statistically meaningful update while avoiding premature budget exhaustion.

More precisely, ML-IQAE employs the budget-aware shot allocation rule
\begin{equation}
m_t = \min\{ m_t^{(1)},\, m_t^{(2)} \},
\qquad m_{\min}\le m_t \le m_{\max},
\label{eq:mt_rule}
\end{equation}
with
\begin{subequations}
\begin{align}
m_t^{(1)} &= \bar m(B)\big(1+\alpha t\big), \label{eq:mt_growth}\\[3pt]
m_t^{(2)} &= \frac{B-\sum_{j<t}(2k_j+1)m_j}{(2k_t+1)\,R_t}. \label{eq:mt_budget}
\end{align}
\end{subequations}
where $B$ is the total oracle-call budget, $\bar m(B)=\Theta(B^{1/4})$ is a smooth base shot scale, $\alpha>0$ is a small
growth rate, and $R_t$ is a prescribed lower bound on the remaining number of rounds. The first term increases the nominal
shot count gradually as the inference progresses, while the second term ensures that sufficient budget is retained to
complete the remaining rounds. The bounds $m_{\min}$ and $m_{\max}$ guarantee that each batch yields a statistically
meaningful update while preventing premature budget exhaustion.

In the present implementation we choose
\begin{subequations}
\begin{align}
\bar m(B) &= 220\,B^{1/4}, \label{eq:mt_params_a}\\
\alpha   &= 0.012, \label{eq:mt_params_b}\\
R_t      &= \max\{10,\,28-\min(t,20)\},
\label{eq:mt_params_c}
\end{align}
\end{subequations}
and clip $m_t$ to $m_{\min}\le m_t\le m_{\max}$ with $m_{\min}=50$ and $m_{\max}=1100$. These values provide stable early-stage
exploration, smooth precision refinement at later rounds, and reliable suppression of spurious admissible modes during
disambiguation.

\begin{algorithm}[!tb]
\caption{High-level ML-IQAE iteration with union-of-intervals feasible sets}
\label{alg:iqae_round2}
\begin{algorithmic}[1]
\Require Initial feasible set $\Theta^{(0)}\subset(0,\pi/2)$ represented as a union of disjoint intervals; global failure tolerance $\delta_{\mathrm{tot}}$; oracle-call budget $B$.
\Ensure Final feasible set $\Theta^{(T)}$ (union of intervals) and an estimate of $a=\sin^2\theta$ (or failure).

\State $t\gets 1$.
\While{budget remains and stopping criterion not met}
    \State Choose Grover depth $k_t$ based on the current feasible set $\Theta^{(t-1)}$.
    \State Choose shots $m_t$ subject to remaining budget (cost $\propto (2k_t+1)m_t$).
    \State Run the amplified circuit $m_t$ times at depth $k_t$ and record successes $h_t$.
    \State Allocate per-round risk $\delta_t \gets \delta_{\mathrm{schedule}}(t,\delta_{\mathrm{tot}})$ \Comment{\eref{eq:delta_schedule}}
    \State Define the round-$t$ success probability $p_t(\theta)=\sin^2\!\big((2k_t+1)\theta\big)$.
    \State Build a binomial confidence interval $[p_L^{(t)},p_U^{(t)}]$ for $p_t(\theta)$ (Clopper--Pearson).
    \State Convert $[p_L^{(t)},p_U^{(t)}]$ into a feasible set for $\theta$:
    \[
      \Theta_t \;=\;\{\theta\in(0,\pi/2): p_t(\theta)\in[p_L^{(t)},p_U^{(t)}]\}.
    \]
    \State Update the global feasible set by intersection:
    \[
      \Theta^{(t)} \;\gets\; \Theta^{(t-1)} \cap \Theta_t.
    \]
    \State Represent $\Theta^{(t)}$ as a union of disjoint intervals and apply likelihood-based hypothesis pruning (keep at most $M_{\max}$ components).
    \If{a disambiguation step is scheduled}
        \State Acquire additional low-depth batches (e.g., $k\in\{0,1,2\}$ with larger shot counts) and update $\Theta^{(t)}$ as above.
    \EndIf
    \State Update point estimate: \newline  $\hat\theta_t=\arg\max_{\theta\in\Theta^{(t)}} \ell(\theta;D_t)$ and $\hat a_t=\sin^2\hat\theta_t$.
    \State $t\gets t+1$.
\EndWhile
\State \Return $\Theta^{(T)}$ and $\hat a_T$ (or $a\in[\sin^2(\theta_L^{(T)}),\sin^2(\theta_U^{(T)})]$ if only bounds are desired).
\end{algorithmic}
\end{algorithm}
\subsubsection{Adaptive shot allocation}
\label{sec:adaptive_m}


This adaptive shot-allocation policy concentrates oracle calls at statistically informative stages of the inference,
improves the stability of likelihood-based pruning, and contributes significantly to the observed reduction in oracle
complexity relative to classical Monte Carlo estimation.

\subsubsection{Multi-hypothesis management}
\label{sec:iqae_hyp}

Because $p_k(\theta)=\sin^2((2k+1)\theta)$ is non-injective, the admissible region $\Theta^{(t)}$ may become disconnected
and consist of multiple disjoint subintervals that are simultaneously consistent with the accumulated data. We
therefore represent the confidence-consistent feasible region using a finite hypothesis set
\begin{equation}
\mathcal{H}^{(t)}=\{\Theta^{(t)}_1,\dots,\Theta^{(t)}_{M_t}\},
\label{eq:hyp_set}
\end{equation}
where each $\Theta^{(t)}_m$ is a disjoint admissible interval that serves as a candidate feasible domain for the
maximum-likelihood estimator defined in Section~\ref{sec:mle_iqae}.

When a new batch $(k_{t+1},m_{t+1},h_{t+1})$ is acquired, each hypothesis is updated by intersection,
\begin{equation}
\Theta^{(t+1)}_m=\Theta^{(t)}_m\cap\Theta_{k_{t+1}},
\qquad m=1,\dots,M_t,
\label{eq:hyp_update}
\end{equation}
and empty hypotheses are discarded. To prevent combinatorial growth, only the $M_{\max}$ surviving hypotheses with the
largest attained likelihood values
$\max_{\theta\in\Theta^{(t)}_m}\ell(\theta;D_t)$ are retained, which prioritizes statistically dominant feasible modes
rather than purely geometric interval width.

\subsubsection{Periodic disambiguation}
\label{sec:iqae_disambig}

The oscillatory and non-injective structure of the inverse response
$p_k(\theta)=\sin^2((2k+1)\theta)$
implies that, as higher amplification depths are employed, the admissible parameter set may fracture into multiple
disconnected components that are simultaneously compatible with the accumulated interval constraints. Many of these
components correspond to alias modes introduced by the periodicity of the sine function rather than to statistically
supported parameter values.



To eliminate such spurious modes, the ML-IQAE framework periodically (e.g., after every 5 rounds) re-estimates low-depth response probabilities
using a significantly increased number of shots over a fixed low-depth set
\[
\mathcal{K}_{\mathrm{dis}}=\{0,1,2\}.
\]
These high-fidelity low-frequency measurements tighten the corresponding confidence intervals and impose strong global
constraints on $\theta$ that rapidly invalidate alias components that are only weakly consistent with the accumulated
data.

For each active hypothesis $\Theta_m^{(t)}$, the refined admissible region is obtained by intersecting with the
high-precision disambiguation constraints,
\[
\Theta^{(t+1)}_{m,k}=\Theta^{(t)}_m \cap \Theta_k, \qquad k\in\mathcal{K}_{\mathrm{dis}},
\]
followed by likelihood-based pruning of the resulting hypothesis set as described in
Section~\ref{sec:iqae_hyp}. This periodic disambiguation mechanism is a key component of the proposed ML-IQAE framework
and is not present in classical IQAE formulations \cite{Grinko2021IQAE}.
\begin{figure}[b]
  \centering
  \includegraphics[width=0.92\linewidth]{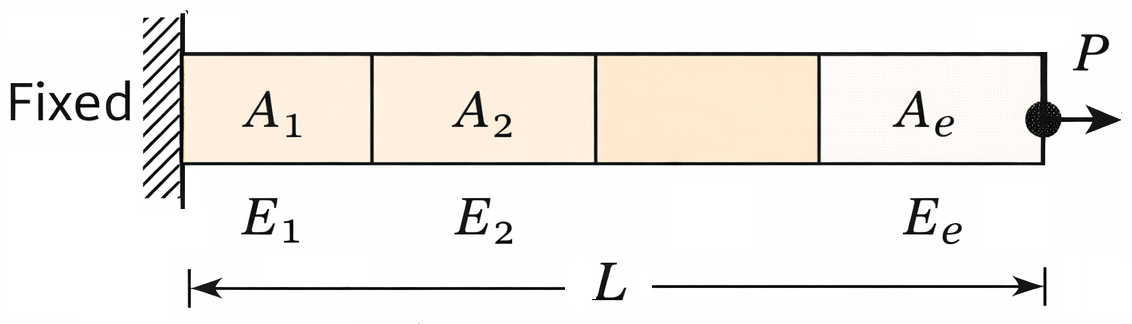}
  \caption{1D axial bar benchmark. Left end fixed ($u(0)=0$) and point load $P$ applied at $x=L$. The bar is discretized into linear elements with constant properties.}
  \label{fig:bar1d_geometry}
\end{figure}\begin{figure*}[tb]
  \centering
  \begin{subfigure}[bt]{0.49\linewidth}
    \centering
    \includegraphics[width=\linewidth]{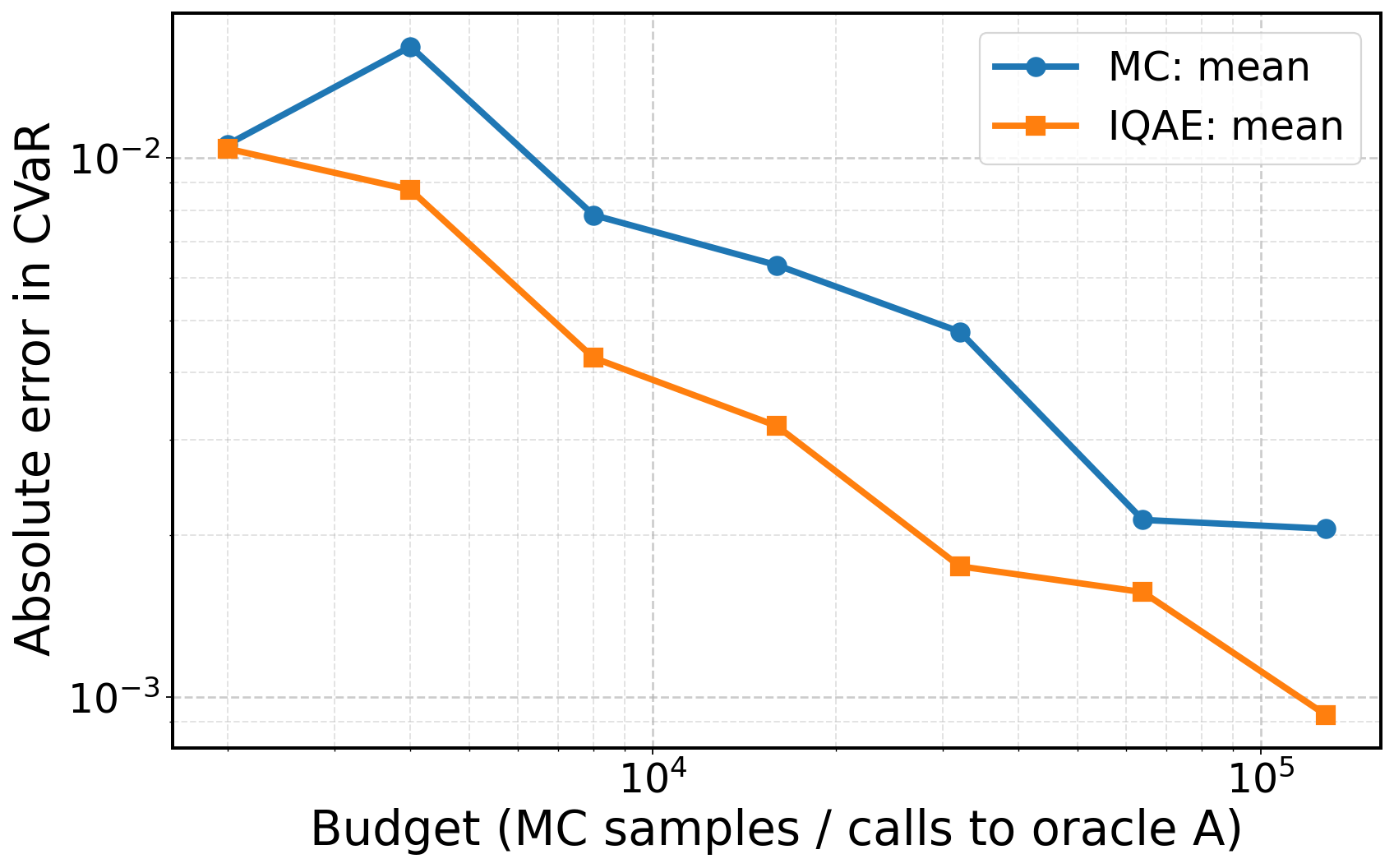}
    \caption{Mean absolute error.}
    \label{fig:bar1d_compliance_mean}
  \end{subfigure}\hfill
  \begin{subfigure}[tb]{0.49\linewidth}
    \centering
    \includegraphics[width=\linewidth]{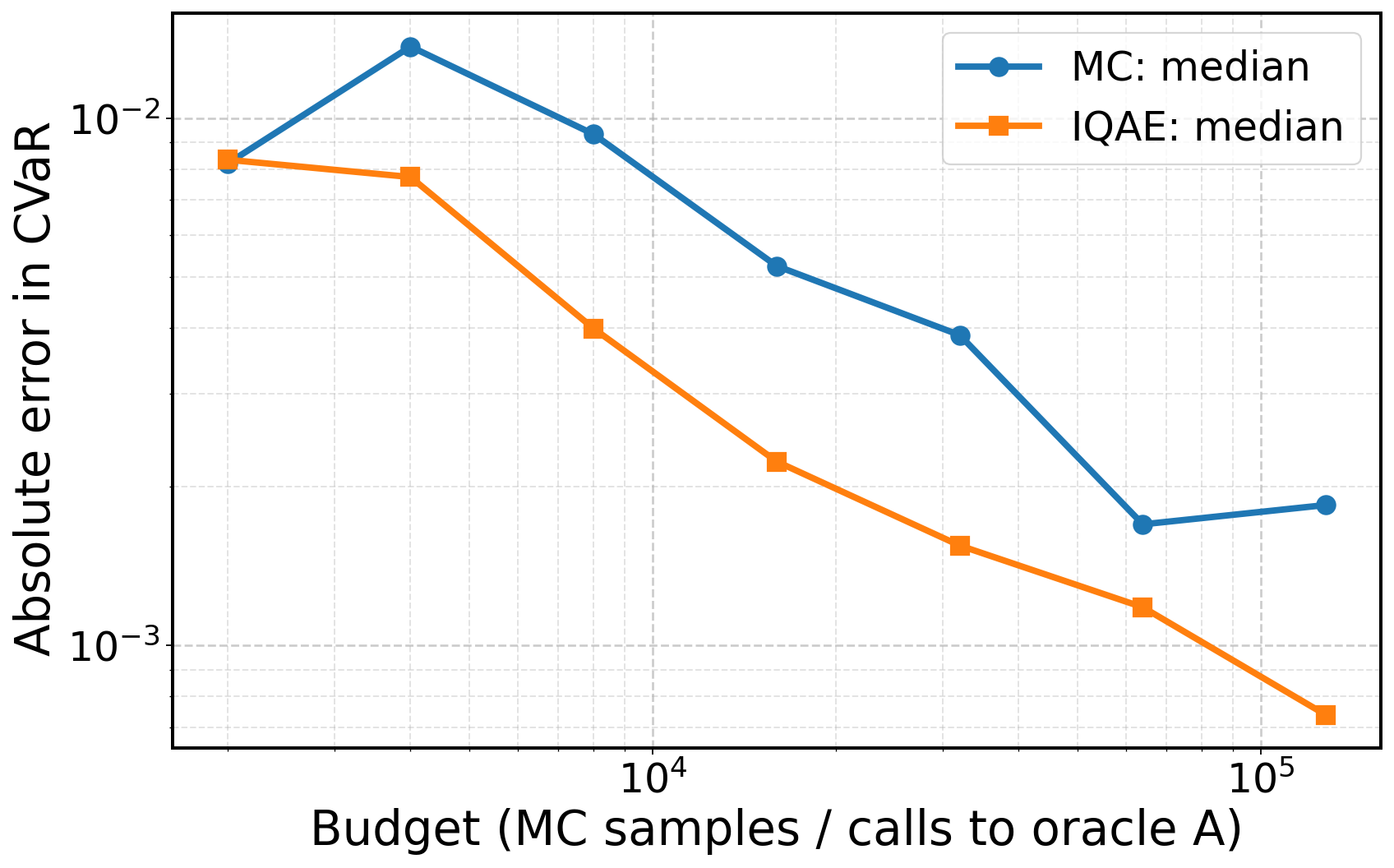}
    \caption{Median absolute error.}
    \label{fig:bar1d_compliance_median}
  \end{subfigure}
  \caption{Absolute error in $\mathrm{CVaR}_{\alpha}(Q)$ for compliance $Q=C$  of the bar benchmark versus requested compute budget.}
  \label{fig:bar1d_compliance_results}
\end{figure*}

To ensure robustness, a bounded restart-on-empty mechanism is employed. If an update causes the feasible parameter set to collapse due to finite-shot noise or aliasing, the most recent observation is discarded and the inference state is reconstructed from earlier consistent measurements. A small number of additional low-depth samples are then used to reinitialize the admissible region. This process is strictly bounded and guarantees recovery without biasing the final estimate.

The complete stabilized ML-IQAE procedure, including likelihood-based hypothesis tracking and periodic disambiguation
is summarized in Algorithm 2. An example of the oracle construction and circuit implementation are provided in Appendix A.

\section{Numerical Benchmarks}
\label{sec:benchmarks}

This section presents a sequence of numerical benchmarks designed to assess the performance of the proposed stabilized ML-IQAE framework for CVaR estimation in stochastic structural mechanics. The examples progress from a one-dimensional axial bar with discrete material uncertainty to two-dimensional cantilever and L-bracket problems with spatially correlated random fields. This hierarchy allows us to isolate estimator behavior from geometric complexity and progressively examine performance under increasing mechanical and statistical challenges.


\subsection{1D axial bar benchmark (discrete random field)}

\begin{table}[t]
\centering
\caption{Standard deviation of the absolute $\mathrm{CVaR}_{\alpha}$ error across repeated runs
for the 1D axial bar problem with compliance as the quantity of interest.}
\label{tab:bar1d_cvar_compliance_std}
\begin{tabular}{c|cc}
\hline
Budget & MC Std & IQAE Std \\
\hline
2000   & 1.03e$-$2 & 6.40e$-$3 \\
4000   & 1.19e$-$2 & 4.92e$-$3 \\
8000   & 4.80e$-$3 & 3.07e$-$3 \\
16000  & 4.16e$-$3 & 2.55e$-$3 \\
32000  & 2.77e$-$3 & 1.31e$-$3 \\
64000  & 2.01e$-$3 & 2.10e$-$3 \\
128000 & 1.50e$-$3 & 8.94e$-$4 \\
\hline
\end{tabular}
\end{table}

To isolate estimator behavior from geometric complexity, we consider a one-dimensional axial bar of length $L$ discretized into $n_e=20$ linear finite elements. The left end of the bar is fixed, $u(0)=0$, and a point load $P$ is applied at the right end, $x=L$ (Figure~\ref{fig:bar1d_geometry}). This configuration admits a closed and interpretable relationship between the response and the applied load, making it a convenient baseline for assessing risk-aware estimators.

Material uncertainty is introduced through a discrete random field defined element-wise. Each element is assigned a Young’s modulus $E_e$ drawn independently from a finite set of 15 admissible levels, resulting in a collection of stochastic realizations with associated probabilities $\{p_i\}$. For this benchmark, we evaluate tail risk using the conditional value-at-risk $\mathrm{CVaR}_{\alpha}$ at confidence level $\alpha=0.95$.

We consider compliance as the quantity of interest, defined as
\[
Q = C = P\,u(L),
\]
where $u(L)$ denotes the tip displacement. For a single point load, compliance is proportional to the tip displacement, and therefore both quantities induce identical stochastic orderings and risk measures up to a constant scaling factor. For this reason, we report results only for compliance, noting that the same conclusions apply to the tip displacement.

Figure~\ref{fig:bar1d_compliance_results} reports the absolute error in $\mathrm{CVaR}_{\alpha}(Q)$ as a function of the requested compute budget, interpreted as either the number of Monte Carlo samples or the number of calls to oracle $A$ for the IQAE-based estimator. The mean absolute error curves reflect average estimator performance across repeated runs, while the median curves provide a more robust characterization of typical behavior by reducing sensitivity to rare but high-error realizations.

At low budgets, MC and IQAE exhibit comparable mean errors, although MC displays larger fluctuations due to sampling variability. As the budget increases, IQAE consistently achieves lower errors and exhibits a steeper decay rate. This improvement is particularly pronounced in the median error curves, where IQAE maintains substantially smaller errors across the full range of budgets, indicating more reliable convergence and fewer extreme outliers.

The variability across runs is quantified by the standard deviation of the absolute CVaR error, reported in Table~\ref{tab:bar1d_cvar_compliance_std}. In this context, the standard deviation measures the spread of estimation errors across independent repetitions at a fixed budget and thus characterizes estimator stability rather than bias. Larger standard deviations indicate a higher likelihood of sporadic large errors, even if the mean performance appears acceptable.
Consistent with the error curves, MC exhibits larger standard deviations, reflecting sensitivity to tail events and sampling noise. IQAE generally maintains smaller standard deviations, particularly at moderate and high budgets, indicating a more concentrated error distribution.
Overall, this simple 1D benchmark confirms that IQAE provides not only improved average accuracy but also enhanced robustness in estimating tail risk.

\subsection{Cantilever Beam Benchmark}
\label{sec:cantilever}

\begin{figure}[t]
    \centering
    \includegraphics[width=1.0\linewidth]{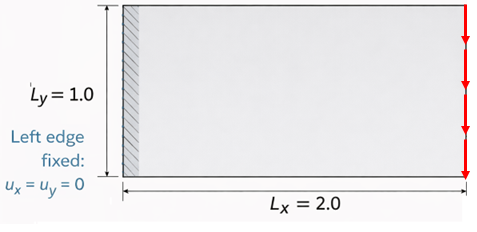}
    \caption{Two-dimensional cantilever beam benchmark. The beam has length $L_x = 2.0$ and height $L_y = 1.0$. The left edge is fully clamped, and a vertical traction load is applied along the right edge.}
    \label{fig:cantilever_geometry}
\end{figure}
\begin{figure*}[b]
    \centering
    \begin{subfigure}[t]{0.48\linewidth}
        \centering
        \includegraphics[width=\linewidth]{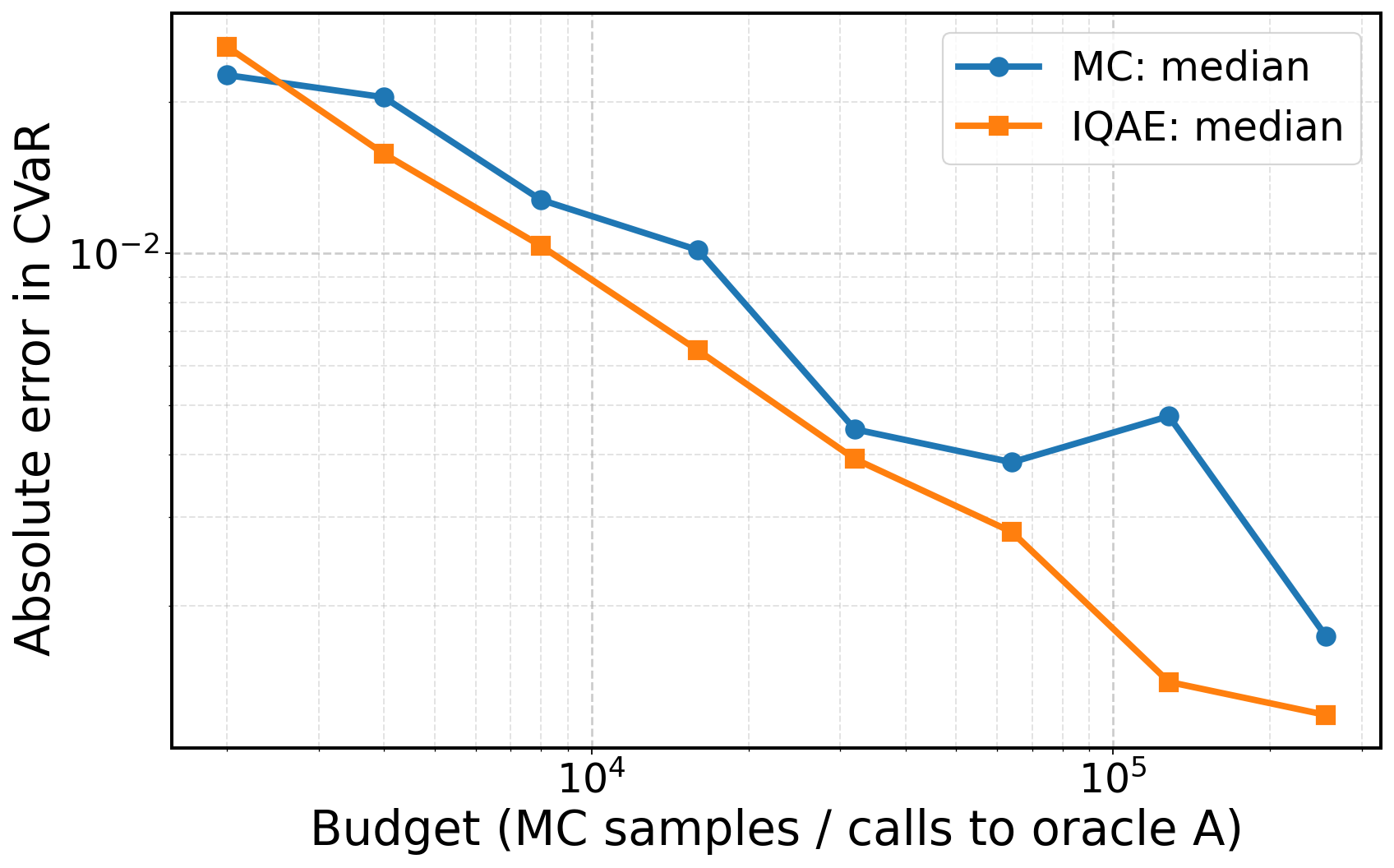}
        \caption{Median absolute CVaR error (compliance).}
        \label{fig:cant_comp_median}
    \end{subfigure}
    \hfill
    \begin{subfigure}[t]{0.48\linewidth}
        \centering
        \includegraphics[width=\linewidth]{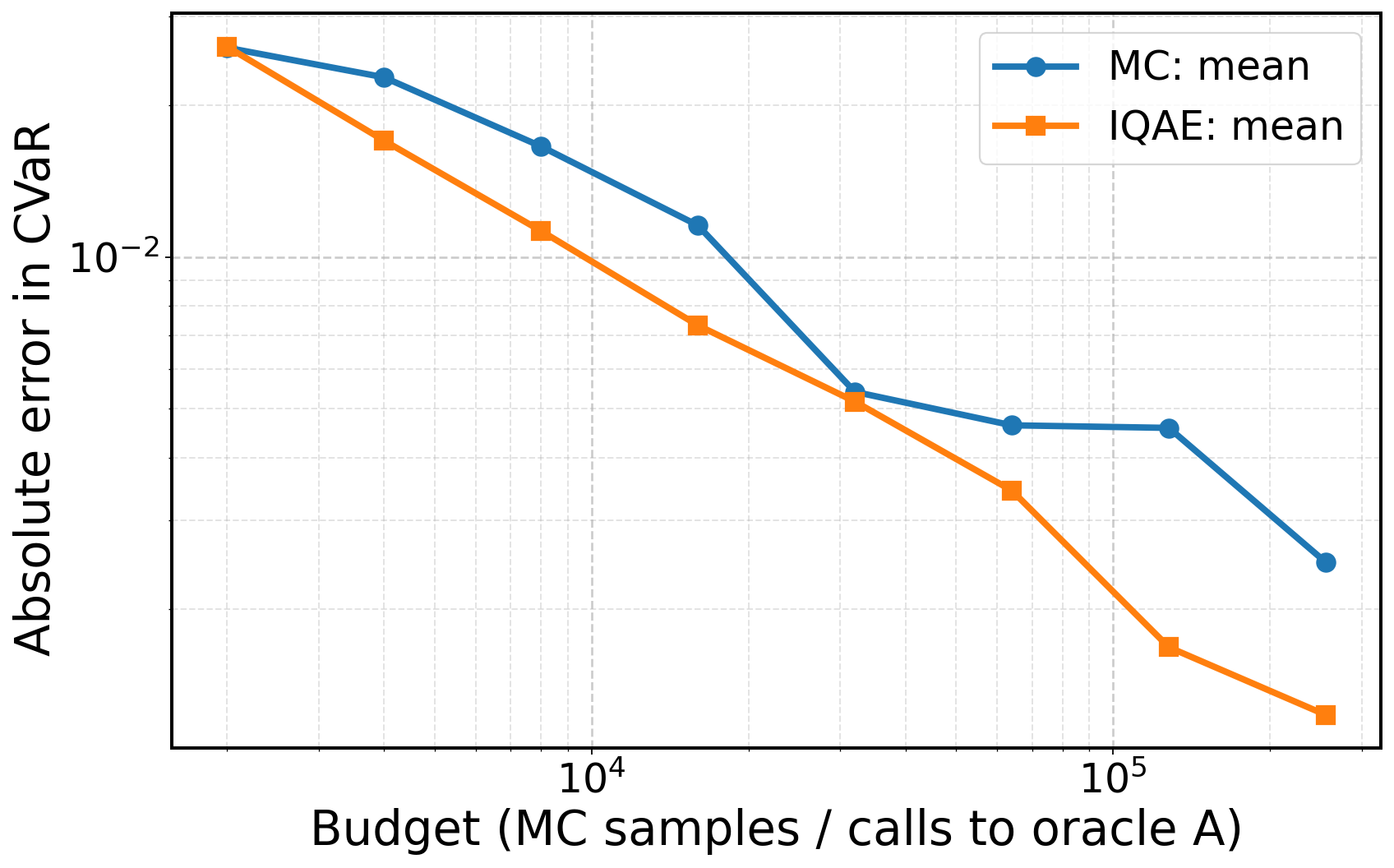}
        \caption{Mean absolute CVaR error (compliance).}
        \label{fig:cant_comp_mean}
    \end{subfigure}
    \caption{Absolute error in $\mathrm{CVaR}_{\alpha}$ estimation of structural compliance for the cantilever beam benchmark as a function of the computational budget. }
    \label{fig:cantilever_compliance}
\end{figure*}

\begin{figure*}[b]
    \centering
    \begin{subfigure}[tb]{0.48\linewidth}
        \centering
        \includegraphics[width=\linewidth]{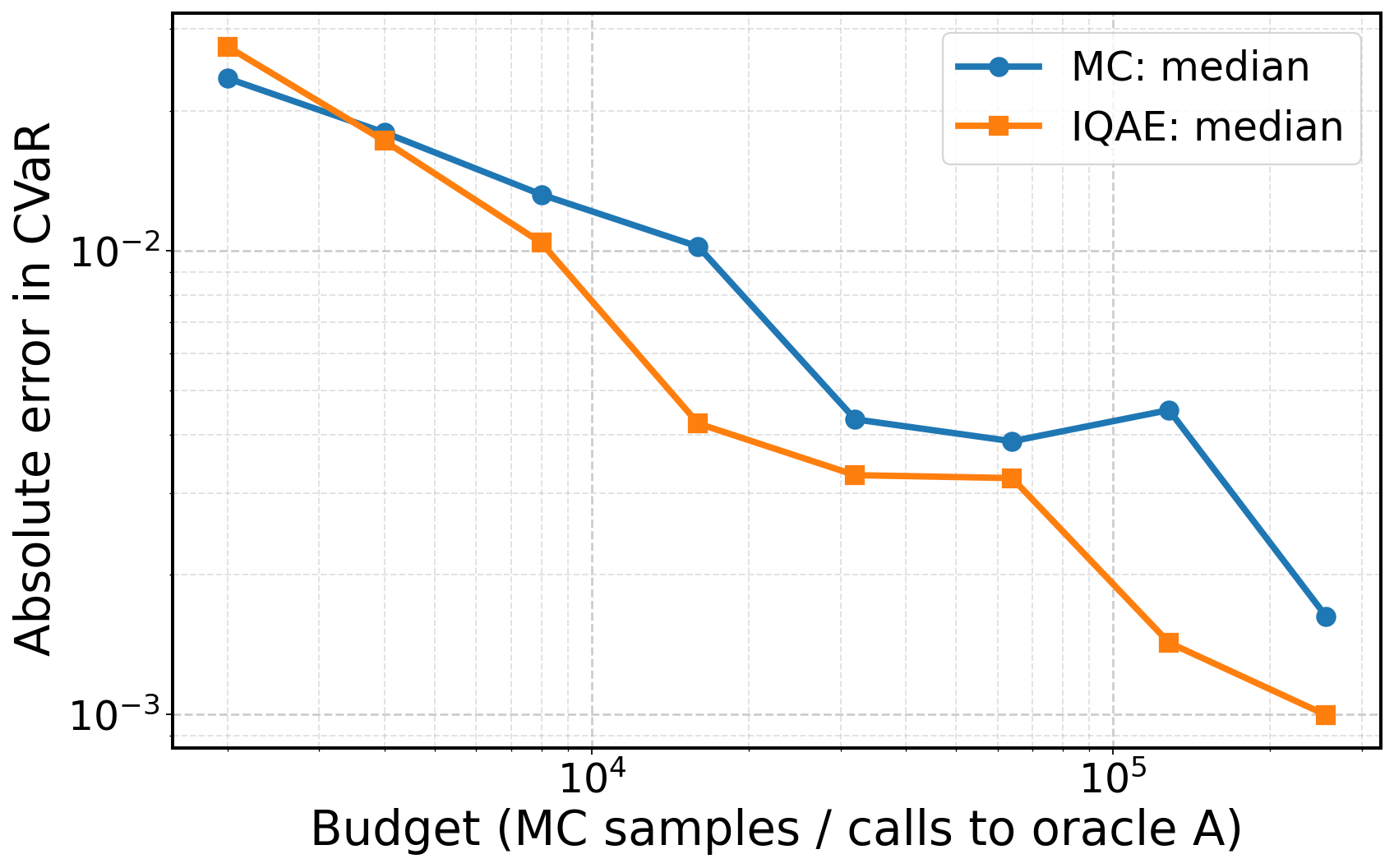}
        \caption{Median absolute CVaR error (tip displacement).}
        \label{fig:cant_disp_median}
    \end{subfigure}
    \hfill
    \begin{subfigure}[tb]{0.48\linewidth}
        \centering
        \includegraphics[width=\linewidth]{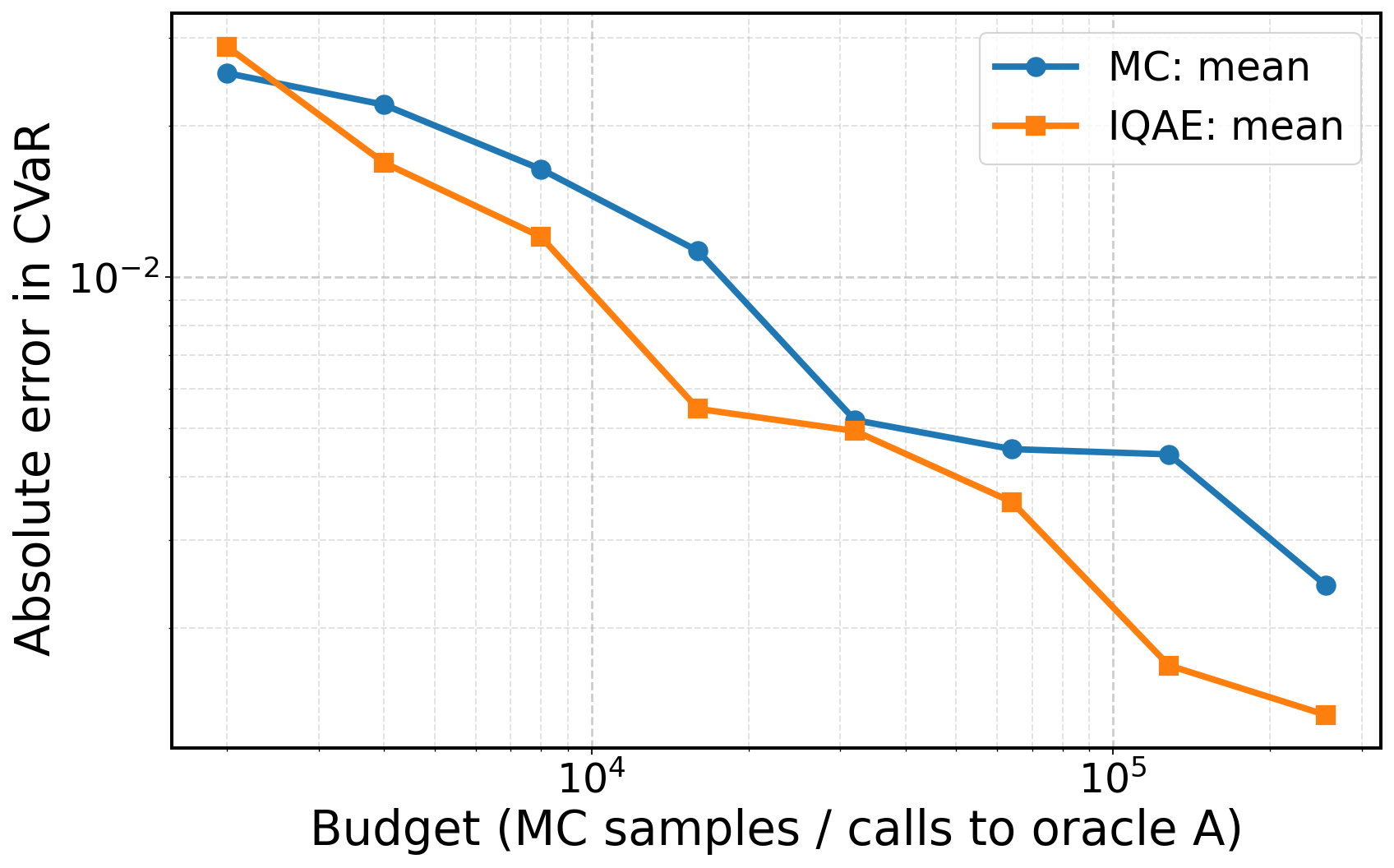}
        \caption{Mean absolute CVaR error (tip displacement).}
        \label{fig:cant_disp_mean}
    \end{subfigure}
    \caption{Absolute error in $\mathrm{CVaR}_{\alpha}$ estimation of vertical tip displacement for the cantilever beam benchmark as a function of the computational budget.}
    \label{fig:cantilever_tipdisp}
\end{figure*}

We next consider a two-dimensional cantilever beam benchmark under plane stress conditions. The geometry, boundary conditions, and loading configuration are shown in Fig.~\ref{fig:cantilever_geometry}. The beam is discretized using bilinear quadrilateral (Q4) finite elements. The left boundary is fully clamped, enforcing zero displacement in both spatial directions, while a uniform vertical traction load is applied along the right edge. 

Material uncertainty is introduced through a spatially correlated random field for the Young’s modulus. The correlation structure is defined by a Gaussian-process prior and implemented using a low-rank Nyström approximation, as described in \sref{nystrom}. This construction enforces smooth spatial variability in material stiffness across the domain while enabling efficient sampling of stochastic realizations.
 The continuous random field is discretized into a finite ensemble of stochastic realizations with associated probabilities, which are used consistently by both Monte Carlo and IQAE-based estimators.

For this benchmark, we evaluate the conditional value-at-risk $\mathrm{CVaR}_{\alpha}$ at confidence level $\alpha = 0.95$ for three quantities of interest computed from the same finite element solutions: (i) structural compliance, (ii) vertical tip displacement measured at the mid-height of the free edge, and (iii) the maximum von Mises stress over all elements. For each QoI, CVaR is estimated using standard Monte Carlo sampling and the proposed IQAE-based estimator. The absolute estimation error is reported as a function of the requested computational budget, interpreted as the number of MC samples or the number of calls to oracle $A$.

\begin{figure*}[tb]
    \centering
    \begin{subfigure}[tb]{0.48\linewidth}
        \centering
        \includegraphics[width=\linewidth]{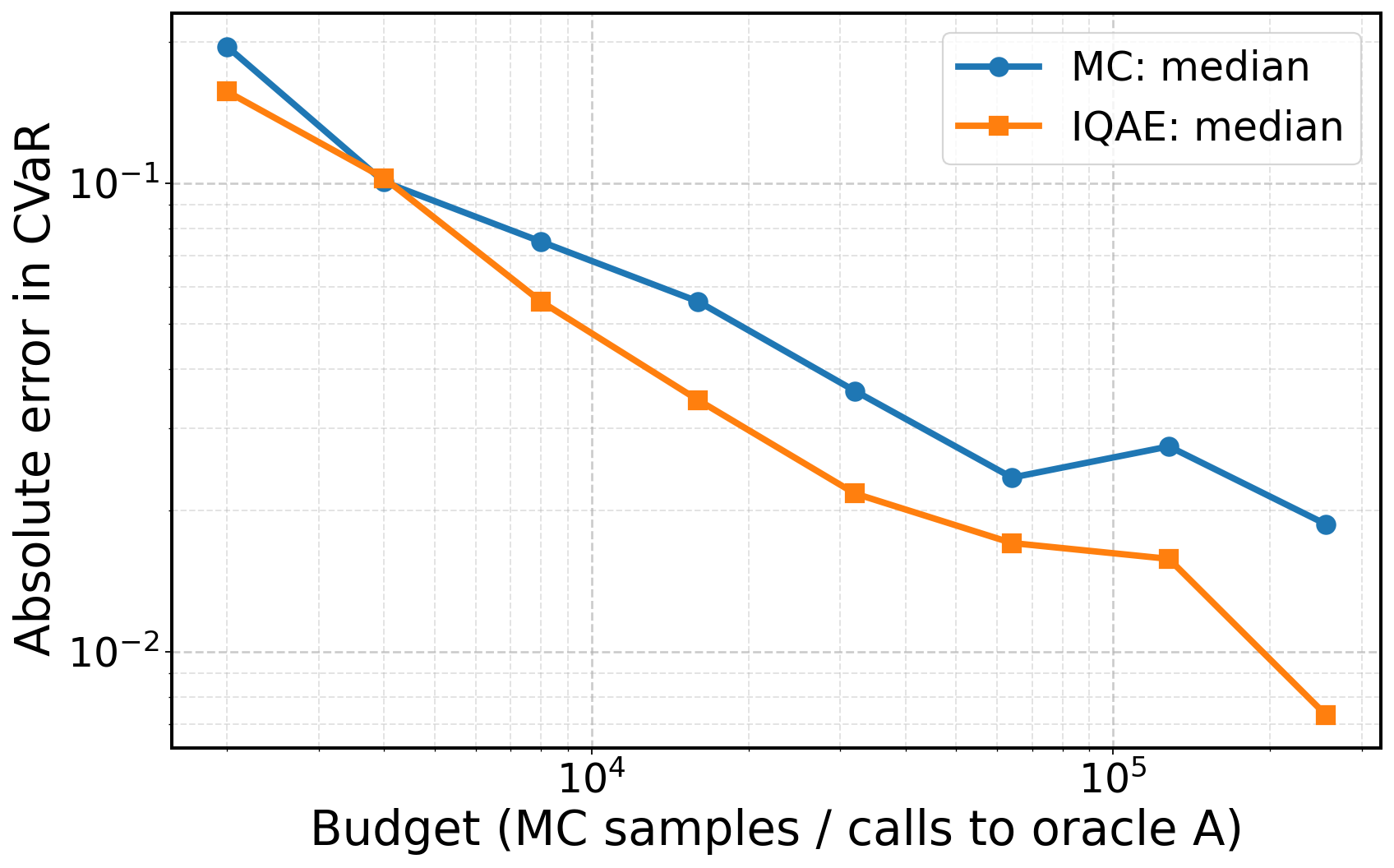}
        \caption{Median absolute CVaR error (max von Mises stress).}
        \label{fig:cant_vm_median}
    \end{subfigure}
    \hfill
    \begin{subfigure}[tb]{0.48\linewidth}
        \centering
        \includegraphics[width=\linewidth]{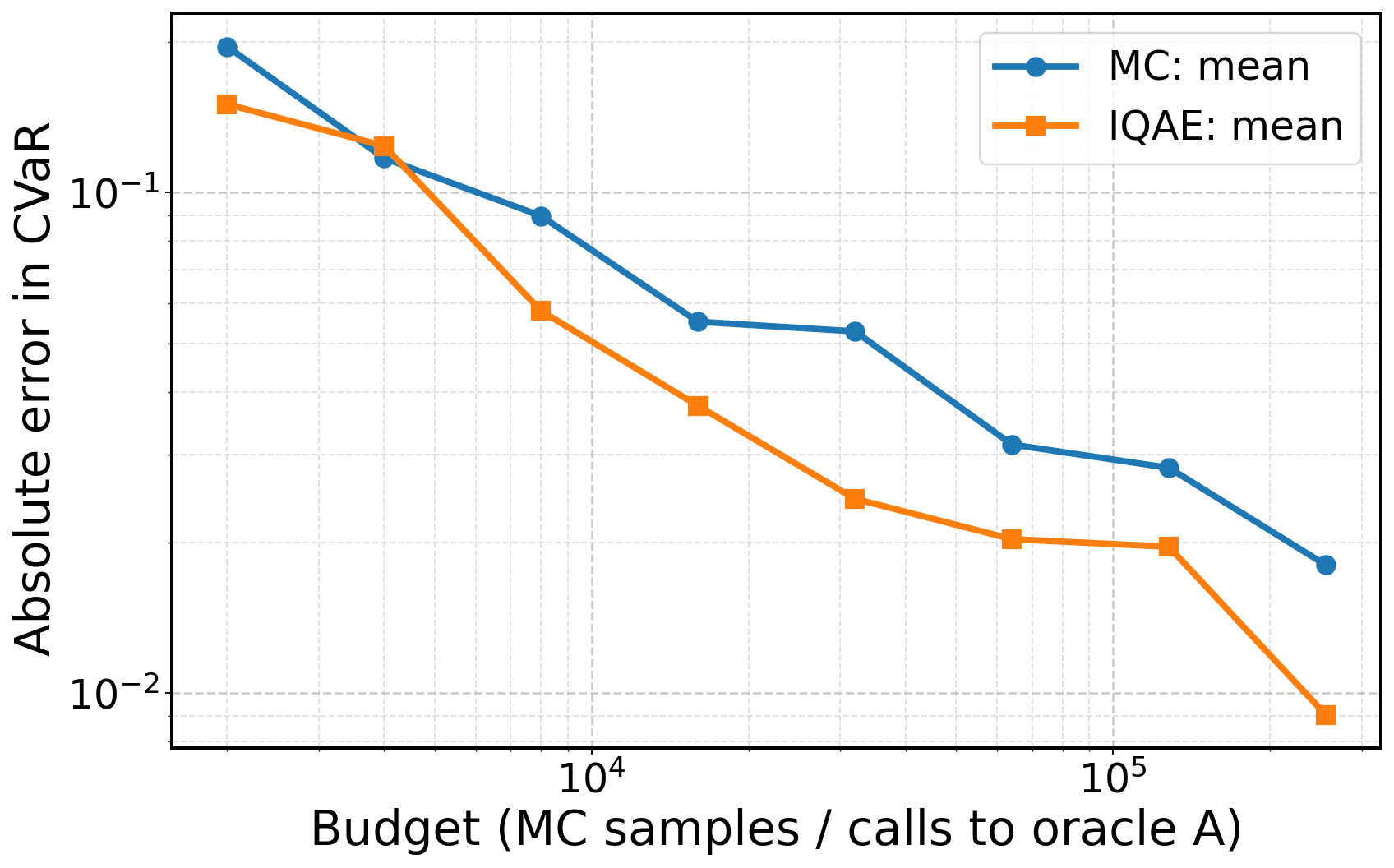}
        \caption{Mean absolute CVaR error (max von Mises stress).}
        \label{fig:cant_vm_mean}
    \end{subfigure}
    \caption{Absolute error in $\mathrm{CVaR}_{\alpha}$ estimation of the maximum von Mises stress for the cantilever beam benchmark as a function of the computational budget.}
    \label{fig:cantilever_stress}
\end{figure*}

Figure~\ref{fig:cantilever_compliance} shows the CVaR estimation error for compliance. Both MC and IQAE exhibit decreasing mean and median errors as the computational budget increases, indicating convergence for both estimators. However, IQAE consistently achieves lower error at comparable budgets. The separation between the median curves is particularly pronounced, highlighting that IQAE yields more reliable performance across repeated runs and is less affected by occasional high-error realizations that inflate the mean error for Monte Carlo sampling.

The corresponding results for the vertical tip displacement are reported in Fig.~\ref{fig:cantilever_tipdisp}. The same qualitative trends are observed: IQAE provides a faster decay of both mean and median CVaR error compared to MC, with the advantage becoming more pronounced at moderate and high budgets. This behavior reflects the improved efficiency of amplitude-estimation-based sampling for displacement-driven quantities, which are sensitive to rare but influential material realizations in the upper tail of the response distribution.

\begin{table*}[t]
\centering
\caption{Standard deviation of the absolute $\mathrm{CVaR}_{\alpha}$ error across repeated runs
for the 2D cantilever problem. Results are shown for Monte Carlo and IQAE estimators
across all three quantities of interest.}
\label{tab:cvar_std_all_qois}
\begin{tabular}{c|cc|cc|cc}
\hline
\multirow{2}{*}{Budget}
& \multicolumn{2}{c|}{Compliance}
& \multicolumn{2}{c|}{Tip displacement}
& \multicolumn{2}{c}{Max von Mises stress} \\
& MC & IQAE & MC & IQAE & MC & IQAE \\
\hline
2000   & 1.82e$-$2 & 2.12e$-$2 & 1.73e$-$2 & 1.72e$-$2 & 1.71e$-$1 & 9.06e$-$2 \\
4000   & 1.43e$-$2 & 9.85e$-$3 & 1.41e$-$2 & 1.05e$-$2 & 1.12e$-$1 & 7.67e$-$2 \\
8000   & 1.24e$-$2 & 8.27e$-$3 & 1.23e$-$2 & 9.48e$-$3 & 6.28e$-$2 & 3.76e$-$2 \\
16000  & 8.13e$-$3 & 5.17e$-$3 & 7.95e$-$3 & 4.21e$-$3 & 3.06e$-$2 & 2.66e$-$2 \\
32000  & 4.95e$-$3 & 4.75e$-$3 & 4.90e$-$3 & 4.22e$-$3 & 4.51e$-$2 & 1.87e$-$2 \\
64000  & 3.56e$-$3 & 2.91e$-$3 & 3.48e$-$3 & 2.57e$-$3 & 2.55e$-$2 & 1.52e$-$2 \\
128000 & 3.31e$-$3 & 1.30e$-$3 & 3.23e$-$3 & 1.35e$-$3 & 1.71e$-$2 & 4.29e$-$2 \\
256000 & 1.87e$-$3 & 1.22e$-$3 & 1.82e$-$3 & 1.17e$-$3 & 1.33e$-$2 & 7.42e$-$3 \\
\hline
\end{tabular}
\end{table*}

Figure~\ref{fig:cantilever_stress} presents the CVaR estimation error for the maximum von Mises stress. Stress-based QoIs exhibit highly skewed and heavy-tailed distributions, making accurate tail estimation particularly challenging for naive Monte Carlo sampling. In this regime, IQAE substantially outperforms MC across all tested budgets, achieving significantly lower mean and median errors. The improvement is especially pronounced at higher budgets, where IQAE continues to reduce error while MC convergence slows due to persistent tail variance.

To quantify estimator robustness, Table~\ref{tab:cvar_std_all_qois} reports the standard deviation of the absolute CVaR error across repeated runs for all three quantities of interest. Across compliance and displacement, IQAE generally exhibits lower or comparable variability relative to MC, with a clear advantage at higher budgets. For the maximum von Mises stress, where response distributions are strongly heavy-tailed, IQAE achieves a substantial reduction in variability over most budgets, underscoring its effectiveness for stress-driven risk metrics.

Overall, the cantilever beam benchmark demonstrates that IQAE delivers consistent accuracy gains and improved robustness over Monte Carlo sampling across compliance-, displacement-, and stress-based CVaR metrics. These improvements are achieved without problem-specific tuning and persist under spatially correlated material uncertainty, highlighting the generality and practical relevance of the proposed approach for risk-aware structural analysis.
\begin{figure}[tb]
    \centering
    \includegraphics[width=\linewidth]{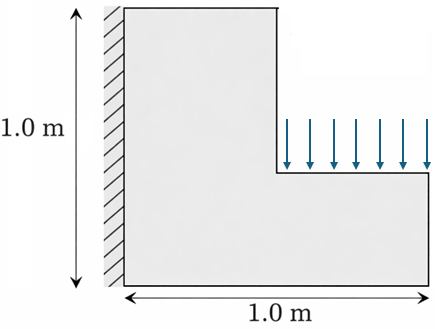}
    \caption{
    Geometry and boundary conditions of the L-shaped benchmark domain.
    }
    \label{fig:lbracket}
\end{figure}

\subsection{L-bracket benchmark: stress-based CVaR under correlated material uncertainty}

The L-bracket benchmark is used to assess the performance of the proposed stabilized ML-IQAE framework
in a mechanically demanding setting characterized by geometric singularities and pronounced stress localization.
As shown in Fig.~\ref{fig:lbracket}, the re-entrant corner of the L-shaped domain induces a classical stress singularity,
resulting in highly concentrated von Mises stress fields near the corner.
This feature makes the upper tail of the stress response distribution particularly sensitive to spatially correlated
material fluctuations and therefore provides a stringent test for stress-based CVaR estimation under limited computational budgets.

\begin{figure*}[!htb]
    \centering
    \begin{subfigure}[t]{0.48\linewidth}
        \centering
        \includegraphics[width=\linewidth]{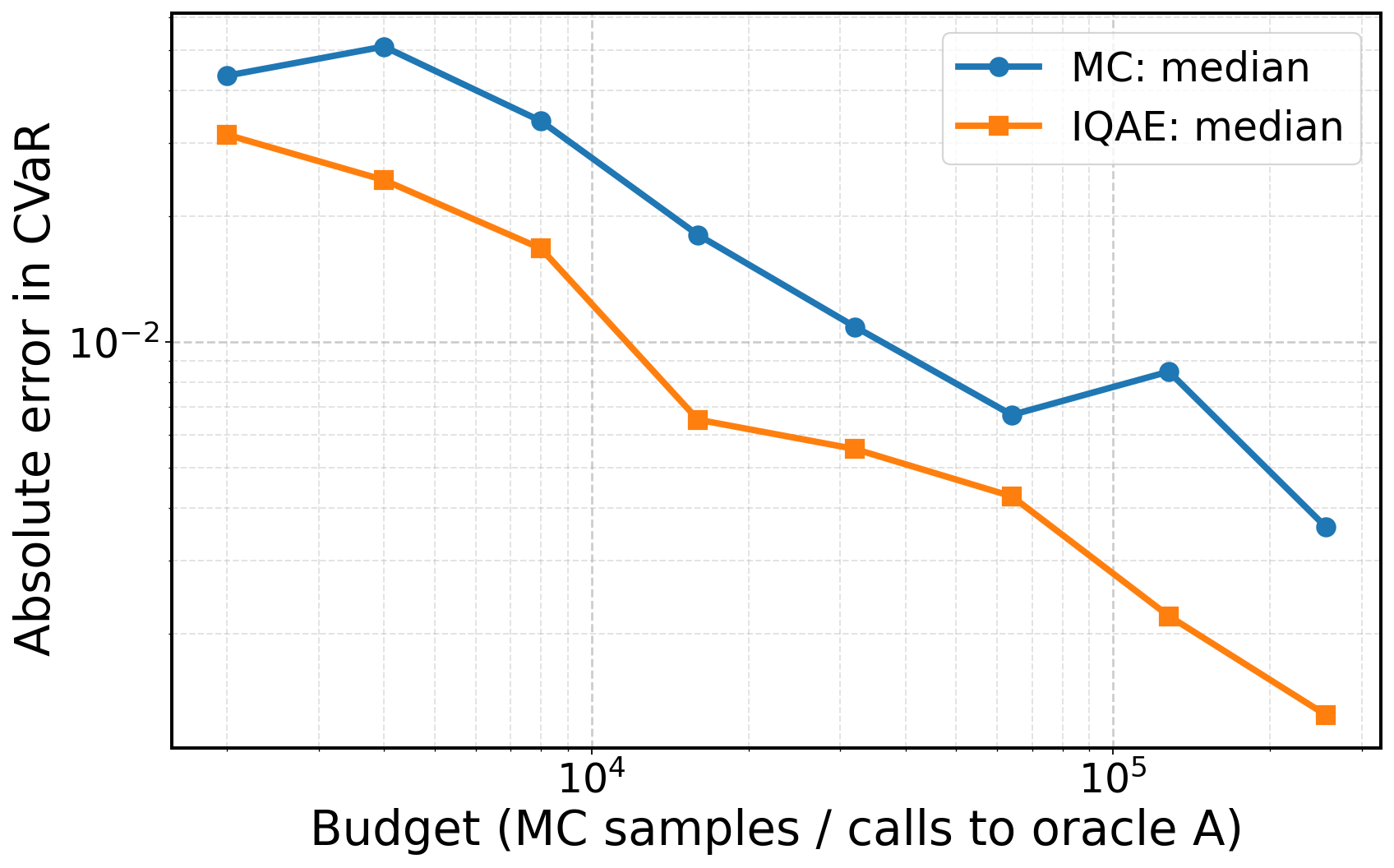}
        \caption{Median absolute error in $\mathrm{CVaR}_{0.95}$.}
        \label{fig:lbracket_median}
    \end{subfigure}
    \hfill
    \begin{subfigure}[t]{0.48\linewidth}
        \centering
        \includegraphics[width=\linewidth]{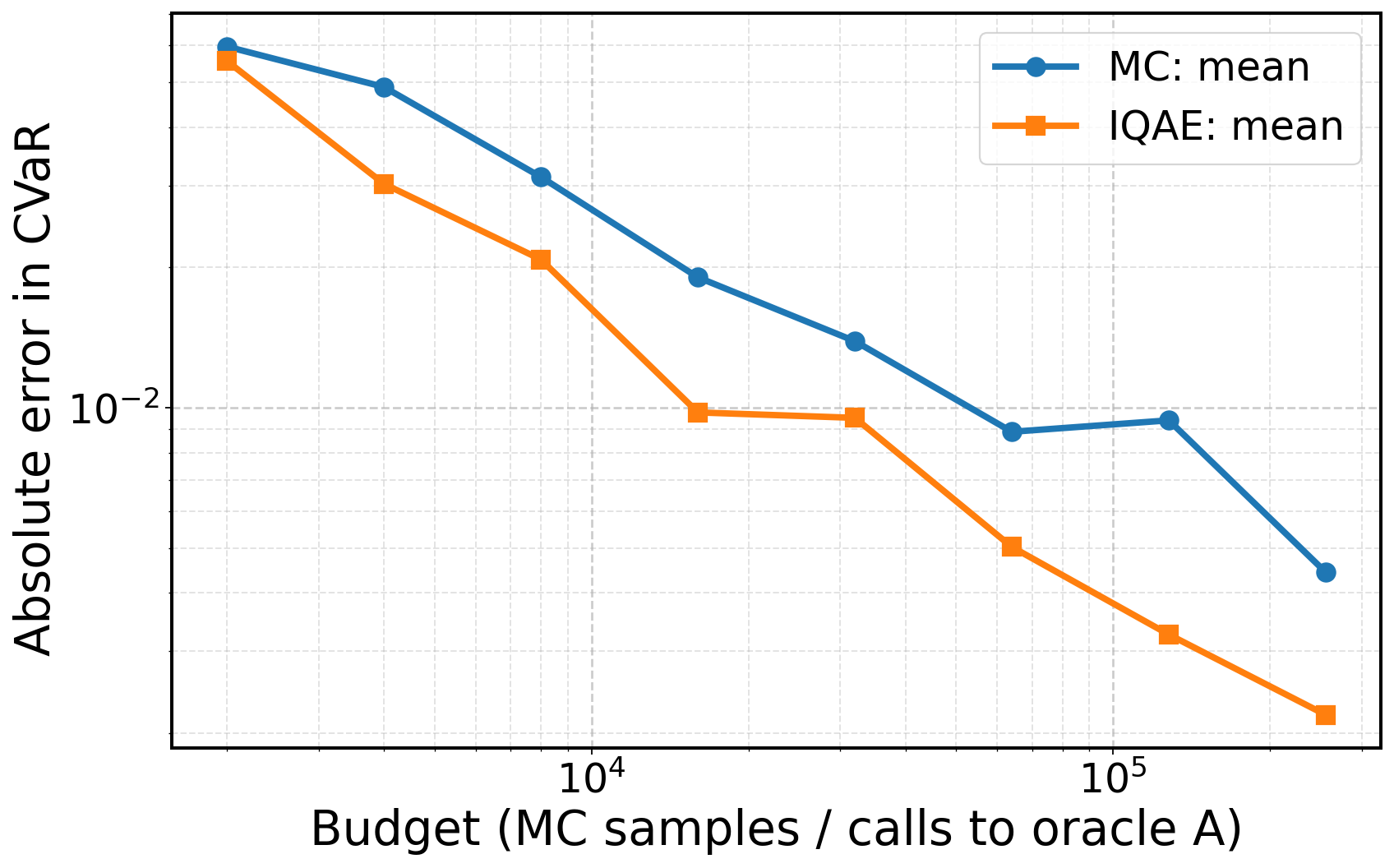}
        \caption{Mean absolute error in $\mathrm{CVaR}_{0.95}$.}
        \label{fig:lbracket_mean}
    \end{subfigure}

    \caption{
    Stress-based CVaR estimation for the L-bracket benchmark under spatially correlated material uncertainty.
    The quantity of interest is the conditional value-at-risk of the maximum von Mises stress over the domain.
    Results compare Monte Carlo sampling with the proposed IQAE-based estimator as a function of the total oracle-call budget.
    }
    \label{fig:lbracket_results}
\end{figure*}

The problem is formulated under plane-stress conditions with uncertainty introduced through a spatially correlated,
lognormal Young’s modulus field.
The same Gaussian-process-based correlation model employed in the cantilever benchmark is used here,
with realizations generated via a Nyström low-rank approximation of the covariance kernel.
A fixed discrete scenario ensemble is constructed, and the conditional value-at-risk of the maximum von Mises stress,
$\mathrm{CVaR}_{0.95}(\sigma_{\mathrm{vm,max}})$, is evaluated.
As in the previous benchmarks, the value-at-risk threshold is computed once from the full scenario distribution
and held fixed throughout the estimation process, so that both Monte Carlo and ML-IQAE estimate the same normalized
hinge expectation. This design isolates the effect of the inference strategy and enables a direct comparison
of oracle efficiency.

Figure~\ref{fig:lbracket_results}(a) reports the median absolute CVaR error as a function of the requested oracle-call budget.
The median is used to characterize typical estimator performance under the highly skewed error distributions
associated with tail-dominated stress responses.
Across all tested budgets, Monte Carlo exhibits slow convergence, while ML-IQAE consistently achieves lower median error
once moderate budgets are available.
The performance gap widens as the budget increases, indicating that ML-IQAE resolves the stress-driven tail behavior
more efficiently than independent sampling.
This improvement is particularly significant in this benchmark, where rare high-stress events dominate the CVaR metric.
\begin{table}[b]
\centering
\caption{Standard deviation of the absolute $\mathrm{CVaR}_{\alpha}$ error across repeated runs
for the L--bracket problem with maximum von Mises stress as the quantity of interest.}
\label{tab:lbracket_cvar_stress_std}
\begin{tabular}{c|cc}
\hline
Budget & MC Std & IQAE Std \\
\hline
2000   & 5.10e$-$2 & 5.19e$-$2 \\
4000   & 3.46e$-$2 & 2.60e$-$2 \\
8000   & 2.23e$-$2 & 2.10e$-$2 \\
16000  & 1.73e$-$2 & 1.49e$-$2 \\
32000  & 1.10e$-$2 & 9.36e$-$3 \\
\hline
\end{tabular}
\end{table}
To complement the median statistics, Fig.~\ref{fig:lbracket_results}(b) shows the mean absolute CVaR error without variance shading.
The mean trends closely mirror the median behavior, confirming that the observed improvement is not driven by a small number
of favorable realizations.
In particular, ML-IQAE maintains a lower mean error than Monte Carlo over the full range of tested budgets and exhibits
a visibly steeper decay rate at higher budgets, consistent with the improved oracle complexity of amplitude-estimation-based inference.

The robustness of the two estimators is quantified in Table~\ref{tab:lbracket_cvar_stress_std}, which reports the standard deviation
of the absolute CVaR error across repeated runs.
Monte Carlo exhibits substantial variability due to the difficulty of consistently sampling rare, high-stress realizations
associated with the re-entrant corner.
In contrast, ML-IQAE demonstrates a systematically reduced standard deviation at moderate and high budgets,
indicating a tighter concentration of estimator outcomes.
This variance reduction reflects the stabilizing effect of likelihood-based hypothesis management and periodic low-depth
disambiguation, which suppress aliasing and finite-shot noise in the amplified response.

At the lowest budgets, ML-IQAE can exhibit variability comparable to Monte Carlo.
This behavior is expected given the oscillatory and non-injective nature of the amplified response function
and the limited statistical information available in this regime.
However, as additional oracle calls are allocated, the admissible parameter region contracts rapidly,
spurious hypotheses are eliminated, and the estimator transitions into a stable convergence regime
with reduced dispersion.

Importantly, no ML-IQAE run failed for this benchmark across any tested budget.
The empirical failure rate is therefore zero, confirming that the bounded restart-on-empty mechanism
is effective even in the presence of severe stress localization and tail sensitivity.
This result underscores the practical robustness of the stabilized inference framework and its suitability
for mechanically realistic CVaR estimation problems.

Overall, the L-bracket results demonstrate that ML-IQAE provides a reliable and sample-efficient alternative
to classical Monte Carlo sampling for stress-based CVaR estimation in geometrically complex structures.
The method achieves lower typical and average errors, exhibits reduced estimator variance,
and maintains strict robustness guarantees without incurring failures,
even under challenging tail-dominated conditions.
\section{Conclusions}

This work presented a stabilized inference framework for risk-aware structural analysis based on iterative quantum amplitude estimation. The proposed ML-IQAE approach targets the efficient and robust estimation of conditional value-at-risk for quantities of interest arising in finite-element models with spatially correlated material uncertainty. By combining likelihood-based hypothesis management, periodic low-depth disambiguation, and a bounded restart-on-empty mechanism, the framework addresses key practical challenges that arise when applying amplitude-estimation-based methods to mechanically realistic problems.

The performance of the proposed method was evaluated across three progressively challenging benchmarks: a one-dimensional axial bar, a two-dimensional cantilever beam, and a stress-dominated L-bracket problem with a re-entrant corner. In all cases, material uncertainty was modeled using spatially correlated random fields constructed from Gaussian-process priors, ensuring physically meaningful variability while preserving computational tractability. For each benchmark, CVaR was evaluated at a fixed confidence level using identical scenario ensembles, enabling a direct comparison between classical Monte Carlo sampling and ML-IQAE under equivalent oracle-call budgets.

Across all benchmarks and quantities of interest, ML-IQAE consistently achieved lower median and mean absolute CVaR errors than Monte Carlo sampling once moderate budgets were available. These gains were most pronounced for stress-based quantities, where heavy-tailed response distributions and localized singularities severely limit the effectiveness of independent sampling. In such regimes, ML-IQAE demonstrated markedly faster error decay and substantially reduced estimator variance, highlighting its ability to focus computational effort on the tail events that dominate CVaR.

In addition to accuracy improvements, the proposed framework exhibited strong robustness properties. Variance across repeated runs was consistently reduced relative to Monte Carlo, particularly at higher budgets, and no inference failures were observed in any benchmark. This confirms that the stabilization mechanisms incorporated into ML-IQAE are effective in suppressing aliasing, finite-shot noise, and spurious hypothesis persistence, even in the presence of highly skewed and non-smooth response distributions.

Overall, the results demonstrate that ML-IQAE provides a reliable and sample-efficient alternative to classical Monte Carlo methods for CVaR estimation in structural mechanics under correlated material uncertainty. The framework is fully compatible with existing finite-element solvers and does not require problem-specific tuning, making it a promising foundation for future quantum-enhanced risk-aware analysis and design workflows. As quantum hardware and fault-tolerant implementations mature, the proposed approach offers a clear pathway for integrating quantum inference primitives into large-scale computational mechanics applications.
\section{Acknowledgments}
The author acknowledges the financial support of the National Science Foundation under Award No. CMMI-2527378.

\bibliographystyle{elsarticle-num}
\bibliography{bibliography.bib}

@book{rockafellar2000optimization,
  title={Optimization of conditional value-at-risk},
  author={Rockafellar, R. Tyrrell and Uryasev, Stanislav},
  journal={Journal of Risk},
  volume={2},
  number={3},
  pages={21--41},
  year={2000},
  publisher={Springer}
}

@article{Acerbi2002CVaR,
  title={On the coherence of expected shortfall},
  author={Acerbi, Carlo and Tasche, Dirk},
  journal={Journal of Banking \& Finance},
  volume={26},
  pages={1487--1503},
  year={2002},
  publisher={Elsevier}
}

@article{rockafellar2002conditional,
  title={Conditional value-at-risk for general loss distributions},
  author={Rockafellar, R. Tyrrell and Uryasev, Stanislav},
  journal={Journal of Banking \& Finance},
  volume={26},
  pages={1443--1471},
  year={2002},
  publisher={Elsevier}
}

@book{UQforEngineers2012,
  title={Uncertainty quantification: theory, implementation, and applications},
  author={Smith, Ralph C},
  year={2024},
  publisher={SIAM}
}

@article{Au2001SubsetSim,
  title={Estimation of small failure probabilities in high dimensions by subset simulation},
  author={Au, S. K. and Beck, J. L.},
  journal={Probabilistic Engineering Mechanics},
  volume={16},
  number={4},
  pages={263--277},
  year={2001},
  publisher={Elsevier}
}

@article{owen2017comparison,
  title={Comparison of surrogate-based uncertainty quantification methods for computationally expensive simulators},
  author={Owen, Nicola E and Challenor, Peter and Menon, Prathyush P and Bennani, Samir},
  journal={SIAM/ASA Journal on Uncertainty Quantification},
  volume={5},
  number={1},
  pages={403--435},
  year={2017},
  publisher={SIAM}
}

@article{xie2023combined,
  title={Combined relevance vector machine technique and subset simulation importance sampling for structural reliability},
  author={Xie, Bin and Peng, Chong and Wang, Yanzhong},
  journal={Applied Mathematical Modelling},
  volume={113},
  pages={129--143},
  year={2023},
  publisher={Elsevier}
}

@book{Ghanem1991KL,
  title={Stochastic finite elements: a spectral approach},
  author={Ghanem, R. and Spanos, P.},
  year={1991},
  publisher={Springer-Verlag}
}

@article{Woerner2019QuantumFinance,
  title={Quantum Risk Analysis},
  author={Woerner, Stefan and Egger, Daniel J.},
  journal={npj Quantum Information},
  volume={5},
  pages={1--12},
  year={2019},
  publisher={Nature Publishing Group}
}

@inproceedings{Airaudo2023UseCVaR,
  title={On the use of risk measures in digital twins to identify weaknesses in structures},
  author={Airaudo, Facundo and Antil, Harbir and Lohner, Rainald and Rakhimov, Umarkhon},
  booktitle={AIAA SCITECH 2024 Forum},
  pages={2622},
  year={2024}
}

@inproceedings{Hong2011MC,
  title={Monte {Carlo} Estimation of Value-at-Risk, Conditional Value-at-Risk and Their Sensitivities},
  author={L. Jeff Hong and Guangwu Liu},
  booktitle={Proceedings of the Winter Simulation Conference (WSC)},
  year={2011}
}

@incollection{Brassard2002QAE,
  title     = {Quantum Amplitude Amplification and Estimation},
  author    = {Brassard, Gilles and H{\o}yer, Peter and Mosca, Michele and Tapp, Alain},
  booktitle = {Quantum Computation and Quantum Information: A Millennium Volume},
  editor    = {Lo, Hoi-Kwong and Spiller, Tim and Popescu, Sandu},
  pages     = {53--74},
  year      = {2002},
  publisher = {American Mathematical Society}
}

@article{Heinrich2002QuantumSummation,
  title   = {Quantum Summation with an Application to Integration},
  author  = {Heinrich, Stefan},
  journal = {Journal of Complexity},
  volume  = {18},
  number  = {1},
  pages   = {1--50},
  year    = {2002},
  publisher = {Elsevier}
}

@article{Montanaro2015MonteCarlo,
  title   = {Quantum Speedup of Monte {Carlo} Methods},
  author  = {Montanaro, Ashley},
  journal = {Proceedings of the Royal Society A},
  volume  = {471},
  number  = {2181},
  pages   = {20150301},
  year    = {2015}
}

@article{Rebentrost2018PRA,
  title   = {Quantum Computational Finance: Monte {Carlo} Pricing of Financial Derivatives},
  author  = {Rebentrost, Patrick and Gupt, Kirti and Bromley, Thomas R.},
  journal = {Physical Review A},
  volume  = {98},
  pages   = {022321},
  year    = {2018},
  publisher = {American Physical Society}
}

@article{Grinko2021IQAE,
  title   = {Iterative Quantum Amplitude Estimation},
  author  = {Grinko, Dmitry and Gacon, Julien and Zoufal, Chris and Woerner, Stefan},
  journal = {npj Quantum Information},
  volume  = {7},
  pages   = {52},
  year    = {2021},
  publisher = {Nature Publishing Group}
}

@article{Suzuki2020QIP,
  title   = {Amplitude Estimation Without Phase Estimation},
  author  = {Suzuki, Yudai and Endo, Suguru and Benjamin, Simon C. and Rieffel, Eleanor G. and McMahon, Peter L. and Yamazaki, Yoshitaka},
  journal = {Quantum Information Processing},
  volume  = {19},
  number  = {2},
  pages   = {75},
  year    = {2020},
  publisher = {Springer}
}

@book{LeMa2010StochasticFEM,
  title     = {Spectral Methods for Uncertainty Quantification: With Applications to Computational Fluid Dynamics},
  author    = {Le Ma{\^\i}tre, Olivier and Knio, Omar},
  publisher = {Springer},
  year      = {2010}
}

@book{Xiu2010UQBook,
  title     = {Numerical Methods for Stochastic Computations: A Spectral Method Approach},
  author    = {Xiu, Dongbin},
  publisher = {Princeton University Press},
  year      = {2010}
}

@book{Melchers1999Reliability,
  title={Structural reliability analysis and prediction},
  author={Melchers, Robert E and Beck, Andr{\'e} T},
  year={2018},
  publisher={John wiley \& sons}
}

@book{Madsen2006Methods,
  title={Methods of structural safety},
  author={Madsen, Henrik O and Krenk, Steen and Lind, Niels Christian},
  year={2006},
  publisher={Courier Corporation}
}

@article{DerKiureghian2005Reliability,
  title={Structural reliability methods for seismic safety assessment: a review},
  author={Der Kiureghian, A},
  journal={Engineering structures},
  volume={18},
  number={6},
  pages={412--424},
  year={1996},
  publisher={Elsevier}
}

@book{Nowak2001Reliability,
  title={Reliability of structures},
  author={Nowak, Andrzej S and Collins, Kevin R},
  year={2012},
  publisher={CRC press}
}

@book{Ang2013Probability,
  title     = {Probability Concepts in Engineering: Emphasis on Applications to Civil and Environmental Engineering},
  author    = {Ang, Alfredo H.-S. and Tang, Wilson H.},
  publisher = {Wiley},
  year      = {2007}
}

@article{Pflug2000VaR,
  title   = {Some Remarks on the Value-at-Risk and the Conditional Value-at-Risk},
  author  = {Pflug, Georg},
  journal = {Probabilistic Engineering Mechanics},
  volume  = {15},
  pages   = {79--86},
  year    = {2000}
}

@article{Blatman2010AdaptivePC,
  title   = {Adaptive sparse polynomial chaos expansion based on least angle regression},
  author  = {Blatman, G. and Sudret, B.},
  journal = {Journal of Computational Physics},
  volume  = {230},
  pages   = {2345--2367},
  year    = {2011}
}

@article{Sudret2008GlobalSA,
  title   = {Global sensitivity analysis using polynomial chaos expansions},
  author  = {Sudret, B.},
  journal = {Reliability Engineering \& System Safety},
  volume  = {93},
  pages   = {964--979},
  year    = {2008}
}

@article{Williams2001Nystrom,
  title   = {Using the Nystr{\"o}m Method to Speed Up Kernel Machines},
  author  = {Williams, Christopher K. I. and Seeger, Matthias},
  journal = {NeurIPS},
  volume  = {13},
  year    = {2001}
}

@book{Rasmussen2006GPML,
  title     = {Gaussian Processes for Machine Learning},
  author    = {Rasmussen, Carl E. and Williams, Christopher K. I.},
  publisher = {MIT Press},
  year      = {2006}
}

@article{Stamatopoulos2019OptionPricing,
  title   = {Option Pricing Using Quantum Computers},
  author  = {Stamatopoulos, Nikos and Egger, Daniel J. and Sun, Yudong and Woerner, Stefan},
  journal = {Quantum},
  volume  = {4},
  pages   = {291},
  year    = {2020}
}

@article{austin2012quantum,
title={Quantum Monte {Carlo} and related approaches},
  author={Austin, Brian M and Zubarev, Dmitry Yu and Lester Jr, William A},
  journal={Chemical reviews},
  volume={112},
  number={1},
  pages={263--288},
  year={2012},
  publisher={ACS Publications}
}

@inproceedings{Grover1996Search,
  title     = {A Fast Quantum Mechanical Algorithm for Database Search},
  author    = {Grover, Lov K.},
  booktitle = {Proceedings of the 28th Annual ACM Symposium on Theory of Computing (STOC)},
  pages     = {212--219},
  year      = {1996},
  publisher = {ACM}
}

@article{Wiebe2016EfficientBayesianPE,
  title   = {Efficient Bayesian Phase Estimation},
  author  = {Wiebe, Nathan and Granade, Christopher},
  journal = {Physical Review Letters},
  volume  = {117},
  number  = {1},
  pages   = {010503},
  year    = {2016}
}

@article{Clopper1934Confidence,
  title   = {The Use of Confidence or Fiducial Limits Illustrated in the Case of the Binomial},
  author  = {Clopper, C. J. and Pearson, E. S.},
  journal = {Biometrika},
  volume  = {26},
  number  = {4},
  pages   = {404--413},
  year    = {1934}
}

@article{Brown2001Interval,
  title   = {Interval Estimation for a Binomial Proportion},
  author  = {Brown, L. D. and Cai, T. T. and DasGupta, A.},
  journal = {Statistical Science},
  volume  = {16},
  number  = {2},
  pages   = {101--133},
  year    = {2001}
}

@article{Montanaro2016FEM,
  title   = {Quantum algorithms and the finite element method},
  author  = {Montanaro, Ashley and Pallister, Sam},
  journal = {Physical Review A},
  volume  = {93},
  number  = {3},
  pages   = {032324},
  year    = {2016},
  doi     = {10.1103/PhysRevA.93.032324}
}

@article{arora2025implementation,
  title={An implementation of the finite element method in hybrid classical/quantum computers},
  author={Arora, Abhishek and Ward, Benjamin M and Oskay, Caglar},
  journal={Finite Elements in Analysis and Design},
  volume={248},
  pages={104354},
  year={2025},
  publisher={Elsevier}
}

@article{Xu2024QCDataDriven,
  title   = {Quantum computing enhanced distance-minimizing data-driven computational mechanics},
  author  = {Xu, Tianyu and Nguyen, Huy and Ortiz, Michael},
  journal = {Computer Methods in Applied Mechanics and Engineering},
  volume  = {418},
  pages   = {116463},
  year    = {2024},
  doi     = {10.1016/j.cma.2023.116463}
}

@article{kim2025variational,
  title={Variational quantum algorithm for constrained topology optimization},
  author={Kim, Jungin E and Wang, Yan},
  journal={Quantum Science and Technology},
  volume={10},
  number={4},
  pages={045025},
  year={2025},
  publisher={IOP Publishing}
}

@article{he2025efficient,
  title={An efficient quantum computing based structural reliability analysis method using quantum amplitude estimation},
  author={He, Jingran},
  journal={Structural Safety},
  volume={114},
  pages={102555},
  year={2025},
  publisher={Elsevier}
}

@book{AbramowitzStegun,
  title     = {Handbook of Mathematical Functions with Formulas, Graphs, and Mathematical Tables},
  author    = {Abramowitz, Milton and Stegun, Irene A.},
  publisher = {National Bureau of Standards},
  series    = {Applied Mathematics Series},
  volume    = {55},
  year      = {1964}
}

@article{NISTDLMF,
  title={NIST digital library of mathematical functions},
  author={Lozier, Daniel W},
  journal={Annals of Mathematics and Artificial Intelligence},
  volume={38},
  number={1},
  pages={105--119},
  year={2003},
  publisher={Springer}
}

@book{NielsenChuang2010,
  title     = {Quantum Computation and Quantum Information},
  author    = {Nielsen, Michael A. and Chuang, Isaac L.},
  publisher = {Cambridge University Press},
  year      = {2010}
}

@article{tabarraei2025variational,
  title={Variational quantum latent encoding for topology optimization},
  author={Tabarraei, Alireza},
  journal={Engineering with Computers},
  volume={41},
  number={6},
  pages={4549--4573},
  year={2025},
  publisher={Springer}
}

@article{Bazant1988CrackBand,
  title   = {Crack band theory for fracture of concrete},
  author  = {Ba{\v{z}}ant, Zden{\v{e}}k P. and Oh, Byung H.},
  journal = {Materials and Structures},
  volume  = {16},
  number  = {3},
  pages   = {155--177},
  year    = {1983},
  publisher = {Springer}
}

@article{Simo1992SDM,
  title   = {An analysis of strong discontinuities induced by strain-softening in rate-independent inelastic solids},
  author  = {Simo, J. C. and Oliver, J. and Armero, F.},
  journal = {Computational Mechanics},
  volume  = {12},
  number  = {5},
  pages   = {277--296},
  year    = {1993},
  publisher = {Springer}
}

@article{Ortiz1999Discontinuous,
  title   = {Finite-deformation irreversible cohesive elements for three-dimensional crack-propagation analysis},
  author  = {Ortiz, Michael and Pandolfi, Alberto},
  journal = {International Journal for Numerical Methods in Engineering},
  volume  = {44},
  number  = {9},
  pages   = {1267--1282},
  year    = {1999},
  publisher = {Wiley}
}

@article{tabarraei2013two,
  title={A two-scale strong discontinuity approach for evolution of shear bands under dynamic impact loads},
  author={Tabarraei, Alireza and Song, Jeong-Hoon and Waisman, Haim},
  journal={International Journal for Multiscale Computational Engineering},
  volume={11},
  number={6},
  year={2013},
  publisher={Begel House Inc.}
}

@article{elapolu2021mechanical,
  title={Mechanical and fracture properties of polycrystalline graphene with hydrogenated grain boundaries},
  author={Elapolu, Mohan SR and Tabarraei, Alireza},
  journal={The Journal of Physical Chemistry C},
  volume={125},
  number={20},
  pages={11147--11158},
  year={2021},
  publisher={ACS Publications}
}

@article{lee2023bi,
  title={Bi-fidelity conditional value-at-risk estimation by dimensionally decomposed generalized polynomial chaos expansion},
  author={Lee, Dongjin and Kramer, Boris},
  journal={Structural and Multidisciplinary Optimization},
  volume={66},
  number={2},
  pages={33},
  year={2023},
  publisher={Springer}
}
\appendix
\section{Explicit circuit-level construction of the oracle and Grover iterates:
a two-qubit worked example}
 \label{app:oracleA_circuit}
 This appendix provides a complete circuit-level realization of the oracle $A$
and Grover amplification operator $G$ used throughout the paper, together with
a fully worked two-qubit numerical example. The goal is to make the abstract
definitions in the main text explicit by showing how scenario-dependent ancilla
rotations are implemented using standard quantum gates, how the oracle-prepared
state $|\Psi\rangle = A|000\rangle$ is constructed step by step, and how successive
Grover iterates act on this state at the level of state vectors. We further
illustrate how amplified measurement outcomes at different Grover depths are
converted into confidence-consistent admissible sets for the unknown rotation
angle $\theta$ using interval-based inference, and how intersections across depths
progressively refine the feasible region. 

\subsection{Scenario-dependent ancilla rotations and block-wise action}
\label{app:oracleA_blocks}

We consider the two-qubit oracle circuit shown in Fig.~\ref{fig:oracleA_circuit}
with index qubits $(q_0,q_1)$ and ancilla $q_2$.
The oracle implements a lookup-table encoding in which each index basis state
applies a prescribed rotation to the ancilla. In this example,
\begin{equation}
\begin{aligned}
|00\rangle &:\quad \text{no rotation}, \\
|01\rangle &:\quad R_y(0.70), \\
|10\rangle &:\quad R_y(1.20), \\
|11\rangle &:\quad R_y(1.80).
\end{aligned}
\label{eq:rotation_table}
\end{equation}

We use the qubit ordering $|q_0 q_1 q_2\rangle$ and the basis
\[
(|000\rangle,|001\rangle,|010\rangle,|011\rangle,
 |100\rangle,|101\rangle,|110\rangle,|111\rangle).
\]

After applying Hadamard gates to the index qubits
\begin{equation}
\begin{aligned}
    |\psi_1\rangle
&=
(H \otimes H \otimes I)|000\rangle \\
&=
\frac12\left(
|000\rangle+|010\rangle+|100\rangle+|110\rangle
\right).
\label{eq:psi1_blocks}
\end{aligned}
\end{equation}

\paragraph{Ancilla rotation conditioned on the $|01\rangle$ index state.}

Because the $|00\rangle$ scenario corresponds to a zero rotation, no circuit block
is required for this component. The first nontrivial encoding therefore targets
the $|01\rangle$ basis state and corresponds to the second block in
Fig.~\ref{fig:oracleA_circuit}. This block consists of an $X$ gate on $q_0$,
followed by a doubly controlled $R_y(0.70)$ rotation on the ancilla, and a second
$X$ gate on $q_0$ to uncompute the pattern conversion. Together, these operations
implement a control-on-$|0\rangle$ condition for $q_0$ and a control-on-$|1\rangle$
condition for $q_1$, ensuring that the rotation is applied exclusively to the
$|01\rangle$ component of the index register.

We first apply an $X$ gate on $q_0$ to convert the control-on-$|0\rangle$ condition
into a standard control-on-$|1\rangle$ form. The resulting state is
\begin{equation}
|\psi_2\rangle
=
(X \otimes I \otimes I)|\psi_1\rangle
=
\frac12\left(
|000\rangle+|010\rangle+|100\rangle+|110\rangle
\right),
\label{eq:psi2_blocks}
\end{equation}
which differs from $|\psi_1\rangle$ only by a permutation of basis labels, leaving
all amplitudes unchanged.

Next, a doubly controlled $R_y(0.70)$ gate is applied to the ancilla. This operation
acts only on the component with $(q_0,q_1)=(1,1)$. Using
\begin{equation}
R_y(\theta)|0\rangle
=
\cos(\theta/2)|0\rangle
+
\sin(\theta/2)|1\rangle,
\label{eq:ry_action_blocks}
\end{equation}
the resulting state is
\begin{equation}
\begin{aligned}
|\psi_3\rangle
=&\;
\frac12|000\rangle
+\frac12|010\rangle
+\frac12|100\rangle \\
&+
\frac12\cos(0.35)|110\rangle
+\frac12\sin(0.35)|111\rangle .
\end{aligned}
\label{eq:psi3_blocks}
\end{equation}

Finally, the $X$ gate on $q_0$ is applied again to uncompute the pattern conversion.
This restores the original index labeling while preserving the ancilla rotation,
yielding
\begin{equation}
\begin{aligned}
|\psi_4\rangle
=&\;
\frac12|000\rangle
+\frac12\cos(0.35)|010\rangle
+\frac12\sin(0.35)|011\rangle \\
&+
\frac12|100\rangle
+\frac12|110\rangle .
\end{aligned}
\label{eq:psi4_blocks}
\end{equation}

The corresponding state vector representation is
\begin{equation}
\vec{\psi}_4
=
\begin{bmatrix}
\frac12 \\
0 \\
\frac12\cos(0.35) \\
\frac12\sin(0.35) \\
\frac12 \\
0 \\
\frac12 \\
0
\end{bmatrix}.
\label{eq:psi4_vector}
\end{equation}

\paragraph{Ancilla rotations for the $|10\rangle$ and $|11\rangle$ index states.}

The Third and fourth blocks in Fig.~\ref{fig:oracleA_circuit} repeat the same
structure to encode the remaining scenarios. Specifically,
\begin{itemize}
\item the $|10\rangle$ component is rotated by $R_y(1.20)$, and
\item the $|11\rangle$ component is rotated by $R_y(1.80)$.
\end{itemize}
No other components are modified.

After all three blocks are applied, the oracle produces
\begin{equation}
\begin{aligned}
|\Psi\rangle = A|000\rangle
&=\frac12\Big(
|00\rangle|0\rangle \\
&+|01\rangle\!\left[\cos(0.35)|0\rangle+\sin(0.35)|1\rangle\right] \\
&+|10\rangle\!\left[\cos(0.60)|0\rangle+\sin(0.60)|1\rangle\right] \\
&+|11\rangle\!\left[\cos(0.90)|0\rangle+\sin(0.90)|1\rangle\right]
\Big).
\end{aligned}
\label{eq:final_oracle_state}
\end{equation}

In vector form,
\begin{equation}
\vec{\Psi}
=
\begin{bmatrix}
\frac12 \\
0 \\
\frac12\cos(0.35) \\
\frac12\sin(0.35) \\
\frac12\cos(0.60) \\
\frac12\sin(0.60) \\
\frac12\cos(0.90) \\
\frac12\sin(0.90)
\end{bmatrix}
\approx
\begin{bmatrix}
0.500000 \\
0.000000 \\
0.469686 \\
0.171449 \\
0.412668 \\
0.282321 \\
0.310805 \\
0.391663
\end{bmatrix}.
\label{eq:Psi_vec_start_G}
\end{equation}

Measuring the ancilla qubit therefore yields outcome $|1\rangle$ with
probability $\sum_i p_i g_i$, consistent with the abstract oracle definition.


\subsection{Grover iterate $G$ as a circuit and continued two-qubit numerical example}
\label{app:grover_circuit_example_correct}

The Grover iterate is defined in \eref{eq:grover_op} and given by 
$ G = - A S_0 A^\dagger S_\chi.$

\paragraph{Operation 1: apply $S_\chi$}
Applying $S_\chi$ multiplies all amplitudes with ancilla bit $q_2=1$ by $-1$:
\begin{equation}
\vec{\Psi}^{(\chi)} = S_\chi\,\vec{\Psi}
\approx
\begin{bmatrix}
0.500000 \\
0.000000 \\
0.469686 \\
-0.171449 \\
0.412668 \\
-0.282321 \\
0.310805 \\
-0.391663
\end{bmatrix}.
\label{eq:Psi_after_Schi_correct_appendix}
\end{equation}

\paragraph{Operation 2: apply $A^\dagger$}
Next, uncompute using $A^\dagger$
\begin{equation}
\vec{v} = A^\dagger \vec{\Psi}^{(\chi)}.
\label{eq:v_def_appendix_correct}
\end{equation}
Since $A$ consists of Hadamards on $(q_0,q_1)$ and scenario-controlled $R_y$ rotations on $q_2$,
$A^\dagger$ is implemented by reversing the same circuit and replacing each $R_y(\theta)$ by $R_y(-\theta)$.
For this example, the resulting vector is (numerically, to 6 digits)
\begin{equation}
\vec{v}
\approx
\begin{bmatrix}
\phantom{-}0.474999 \\
-0.637526 \\
\phantom{-}0.206179 \\
\phantom{-}0.171507 \\
\phantom{-}0.407422 \\
\phantom{-}0.315417 \\
-0.088601 \\
\phantom{-}0.150602
\end{bmatrix}.
\label{eq:after_Adag_numeric_appendix_correct}
\end{equation}

\paragraph{Operation 3: apply $S_0$}
The reflection $S_0 = I - 2|000\rangle\langle 000|$ flips the sign of the $|000\rangle$ amplitude only:
\begin{equation}
\vec{v}^{(0)} = S_0\,\vec{v}
\approx
\begin{bmatrix}
-0.474999 \\
-0.637526 \\
\phantom{-}0.206179 \\
\phantom{-}0.171507 \\
\phantom{-}0.407422 \\
\phantom{-}0.315417 \\
-0.088601 \\
\phantom{-}0.150602
\end{bmatrix}.
\label{eq:after_S0_numeric_appendix_correct}
\end{equation}

\paragraph{Operation 4: apply $A$ (complete one Grover iterate)}
Finally,
\begin{equation}
\vec{\Psi}^{(G)} = -A\,\vec{v}^{(0)}.
\label{eq:PsiG_def_appendix_correct}
\end{equation}
Carrying out the multiplication yields
\begin{equation}
\vec{\Psi}^{(G)}
\approx
\begin{bmatrix}
-0.025001 \\
\phantom{-}0.000000 \\
-0.023485 \\
\phantom{-}0.334325 \\
-0.020634 \\
\phantom{-}0.550526 \\
-0.015541 \\
\phantom{-}0.763743
\end{bmatrix}.
\label{eq:after_one_G_numeric_appendix_correct}
\end{equation}

\paragraph{Amplified success probability $p_1$}
Success corresponds to ancilla $q_2=1$, i.e. basis states
$|001\rangle,|011\rangle,|101\rangle,|111\rangle$ (entries 2,4,6,8 in our ordering).
Thus,
\begin{equation}
\begin{aligned}
p_1
&=
|(\vec{\Psi}^{(G)})_{001}|^2
+ |(\vec{\Psi}^{(G)})_{011}|^2 \\
&\quad
+ |(\vec{\Psi}^{(G)})_{101}|^2
+ |(\vec{\Psi}^{(G)})_{111}|^2 \\
&\approx
(0.000000)^2
+ (0.334325)^2\\
&+ (0.550526)^2
+ (0.763743)^2
&\approx 0.998156.
\end{aligned}
\label{eq:p1_numeric_appendix_correct}
\end{equation}

\paragraph{Check against amplitude-amplification theory}
The unamplified success probability is obtained from $|\Psi\rangle=A|000\rangle$:
For the oracle-prepared state $|\Psi\rangle=A|000\rangle$, the unamplified success
probability is
\begin{equation}
\begin{aligned}
a = p_0
&= \frac14\!\left(
\sin^2(0.35)
+ \sin^2(0.60)
+ \sin^2(0.90)
\right) \\
&\approx 0.262500 .
\end{aligned}
\label{eq:p0_numeric}
\end{equation}

Writing $a=\sin^2\theta$ gives $\theta \approx 0.537916$, and the Grover prediction is
\[
p_1 = \sin^2(3\theta) \approx 0.998156,
\]
which matches Eq.~\eqref{eq:p1_numeric_appendix_correct}.


\subsection{Numerical illustration of interval-based inference for $k=0$ and $k=1$}
\label{app:interval_example}




At depth $k=0$, the success probability satisfies
$p_0(\theta)=\sin^2\theta$.
Suppose the amplified circuit is executed $m_0=1000$ times and
$h_0=262$ successes are observed, corresponding to an empirical frequency
$\hat p_0=0.262$.

Using the Clopper--Pearson construction defined in
Section~\ref{sec:binomial_ci}, the resulting confidence interval for $p_0$ is
\begin{equation}
p_0 \in [0.23498,\;0.29043].
\label{eq:p0_interval_numeric}
\end{equation}
Since $\sin^2\theta$ is monotone on $(0,\pi/2)$, this interval maps uniquely to
an admissible interval for $\theta$:
\begin{align}
\Theta_0
&=
\big[
\arcsin(\sqrt{0.23498}),\;
\arcsin(\sqrt{0.29043})
\big]  \nonumber\\ 
&=
[0.50607,\;0.56915]\ \text{rad}.
\label{eq:theta_interval_k0_numeric}
\end{align}


At depth $k=1$, the response function is
$p_1(\theta)=\sin^2(3\theta)$.
Suppose that $m_1=1000$ executions yield $h_1=998$ successes, corresponding to
$\hat p_1=0.998$.
The associated Clopper--Pearson confidence interval is
\begin{equation}
p_1 \in [0.99279,\;0.99976].
\label{eq:p1_interval_numeric}
\end{equation}

Following the interval-to-angle mapping described in
Section~\ref{sec:iqae_core}, this probability interval induces two admissible
branches for $\theta$ on $(0,\pi/2)$:
\begin{equation}
\Theta_1
=
[0.49527,\;0.51841]\ \cup\ [0.52879,\;0.55193]\ \text{rad}.
\label{eq:theta_interval_k1_numeric}
\end{equation}
The presence of two disjoint intervals reflects the non-injective nature of the
mapping $\theta \mapsto \sin^2(3\theta)$.

\subsection{Intersection of admissible sets}

Combining information from depths $k=0$ and $k=1$ corresponds to intersecting
the admissible sets $\Theta_0$ and $\Theta_1$
\begin{equation}
\begin{aligned}
\Theta^{(1)}
&=
\Theta_0 \cap \Theta_1  \\
&=
[0.50607,\;0.51841]\ \cup\ [0.52879,\;0.55193]\ \text{rad}.
\label{eq:theta_intersection_numeric}
\end{aligned}
\end{equation}

Thus, after two rounds, the feasible region consists of two remaining hypotheses.
In the full ML-IQAE procedure, additional Grover depths or scheduled
disambiguation rounds are used to eliminate spurious branches and isolate a
single confidence-consistent interval for $\theta$. Once a unique interval is
identified, bounds or point estimates for the amplitude
$a=\sin^2\theta$ follow immediately.

\end{document}